\documentclass[final,12pt]{clear2025} % Include author names

\usepackage[utf8]{inputenc} % allow utf-8 input
\usepackage[T1]{fontenc}    % use 8-bit T1 fonts
\usepackage{hyperref}       % hyperlinks
\usepackage{booktabs}       % professional-quality tables
\usepackage{amsfonts}       % blackboard math symbols
\usepackage{nicefrac}       % compact symbols for 1/2, etc.
\usepackage{microtype}      % microtypography
\usepackage{xcolor}         % colors

\usepackage{hyperref}
\usepackage{setspace}
\usepackage{booktabs}
\usepackage{wrapfig}
\usepackage{caption}
\usepackage{booktabs}
\usepackage[thinc]{esdiff}

\usepackage{tikz}
\usetikzlibrary{bayesnet}
\usetikzlibrary{arrows}
\usepackage{color}
\usetikzlibrary{backgrounds}

\usepackage{paralist}
\usepackage{mathtools}
\usepackage{bbm}
\usepackage{xcolor}
\usepackage{hyperref}
\usepackage{caption}
\usepackage{subcaption}
\usepackage{enumitem}
\usepackage{wrapfig}

\usepackage{tikz}
\usetikzlibrary{bayesnet}
\usetikzlibrary{arrows}
\usepackage{color}
\usetikzlibrary{backgrounds}

\usetikzlibrary{shapes,arrows,fit,chains,matrix}
\usetikzlibrary{positioning, shapes.geometric}
\tikzstyle{node_no_draw} = [circle, minimum size = 10mm]
\tikzstyle{node} = [circle, minimum size = 10mm, draw =black!80]
\tikzstyle{nodeinter} = [rectangle, minimum size = 10mm, draw =black!80, fill=gray!30]
\tikzstyle{nodeinterwhite} = [rectangle, minimum size = 10mm, draw =black!80]
\tikzstyle{nodeobserved} = [circle, minimum size = 10mm, draw =black!80, fill=gray!30]
\tikzstyle{box} = [rectangle, draw =black!0]
\tikzstyle{arrow} = [thick,->,>=stealth,line width=0.3mm]
\tikzstyle{arrow2} = [dashed,->,>=stealth,line width=0.6mm]

\usetikzlibrary{calc}
\makeatletter
\tikzset{
  prefix after node/.style={
    prefix after command={\pgfextra{#1}}
  },
  /semifill/ang/.store in=\semi@ang,
  /semifill/ang=0,
  semifill/.style={
    circle, draw,
    prefix after node={
      \typeout{aaa \semi@ang}
      \let\nodename\tikz@last@fig@name
      \fill[/semifill/.cd, /semifill/.search also={/tikz}, #1]
        let \p1 = (\nodename.north), \p2 = (\nodename.center) in
        let \n1 = {\y1 - \y2} in
        (\nodename.\semi@ang) arc [radius=\n1, start angle=\semi@ang, delta angle=180];
    },
  }
}

\usepackage{multirow}
\usepackage{booktabs}

\usepackage{algorithm}
\usepackage{algpseudocode}

\usepackage{enumitem}

\title[A Geometric Take on Matching Methods for Treatment Effect Estimation]{Beyond Flatland: A Geometric Take on Matching Methods for Treatment Effect Estimation}
\usepackage{times}
\clearauthor{%
 \Name{Melanie F. Pradier} \Email{melanief@microsoft.com}\\
 \addr Microsoft Research, Cambridge UK
 \AND
 \Name{Javier Gonz\'alez} \Email{gonzalez.javier@microsoft.com}\\
 \addr Microsoft Research, Cambridge UK%
}

\newcommand{\given}{{\, | \,}}

\def \G {\mathbf{G}}

\DeclareMathOperator*{\argmin}{arg\,min}

\definecolor{ao(english)}{rgb}{0.0,0.5, 0.0}
\definecolor{applegreen}{rgb}{0.55,0.71, 0.0}
\definecolor{amethyst}{rgb}{0.6, 0.4,0.8}

\begin{document}

\maketitle

\begin{abstract}%
Matching is a popular approach in causal inference to estimate treatment effects by pairing treated and control units that are most similar in terms of their covariate information. However, classic matching methods completely ignore the geometry of the data manifold, which is crucial to define a meaningful distance for matching, and struggle when covariates are noisy and high-dimensional. In this work, we propose \textit{GeoMatching}, a matching method to estimate treatment effects that takes into account the intrinsic data geometry induced by existing causal mechanisms among the confounding variables. First, we learn a low-dimensional, latent Riemannian manifold that accounts for uncertainty and geometry of the original input data. Second, we estimate treatment effects via matching in the latent space based on the learned latent Riemannian metric. We provide theoretical insights and empirical results in synthetic and real-world scenarios, demonstrating that GeoMatching yields more effective treatment effect estimators, even as we increase input dimensionality, in the presence of outliers, or in semi-supervised scenarios.
% or as maintains stability in the presence of outliers, and can handle semi-supervised scenarios.
% is robust against outliers
%to demonstrate the effectiveness of our method.
\end{abstract}
\vspace{0.1cm}


\section{Introduction}\label{sec:intro}

Causal inference based on observational data plays a pivotal role in unveiling cause-effect relationships, thereby guiding evidence-based practices across numerous sectors including medicine~\citep{rosenbaum_optimal_2012}, public policy~\citep{ben-michael_policy_2023}, epidemiology~\citep{westreich_transportability_2017}, and econometrics~\citep{sekhon_matching_2012}.
% , razonable_casirivimab-imdevimab_2021 -> medicine
The goal is to estimate the causal effect of a \textit{treatment} $T$ on a given \textit{outcome} $Y$ in the presence of (pre-treatment) input covariates or \textit{confounders} $X$, as depicted in Figure~\ref{sec:intro}A.
%
%
%
%
%Causal inference based on observational data plays a pivotal role across numerous sectors including medicine~\citep{rosenbaum_optimal_2012, razonable_casirivimab-imdevimab_2021}, public policy~\citep{ben-michael_policy_2023}, epidemiology~\citep{westreich_transportability_2017}, and econometrics~\citep{sekhon_matching_2012}. The goal is to estimate the causal effect of a treatment $T$ on a given outcome $Y$ given some input (pre-treatment covariates $X$.

Confounders are responsible for the so-called confounding bias~\citep{rubin_causal_2005}, a fundamental issue in observational studies where there exists an imbalance/discrepancy in the distributions of treated and control units, due to the treatment depending on them instead of being randomly assigned to the observed population.
Matching is a well-established approach in causal inference to alleviate confounding bias~\citep{stuart_matching_2010},
%The goal is to estimate the effect of a treatment $T$ on an outcome $Y$
where the goal is to pair suitable control units to each treatment unit (and viceversa) based on the similarity of their covariate information.
%pre-treatment covariates $X$, as shown in Figure~\ref{fig:intro_panel}(a).% which involves the process of pairing treatment units with control units on the basis of their covariate information.
%Matching Method (MM) useful (treatment/control units need to be matched to control/treatment units based on covariate information), shown success in several fields, etc (...) \citep{stuart_matching_2010}.

% O Clivio et.al:  "Matching methods, however, generally suffer from the curse of dimensionality (Abadie and Imbens, 2006a; Roberts et al., 2020), rendering them impractical for many modern highdimensional datasets, such as electronic health records or clinical images."
Traditional matching methods suffer from two main limitations: first, they operate in the space of raw input covariates $\mathcal{X}$ which are noisy and high-dimensional, so it becomes harder to build exact matches~\citep{luo_matching_2017}. Indeed, distances become increasingly meaningless for higher dimensions~\citep{goos_surprising_2001, abadie_large_2006}. Second, observations are assumed to live in Euclidean spaces, overlooking the data manifold's geometry: this may lead to distorted distances~\cite{tosi_metrics_2014}, which result in undesired matched units and imprecise estimates of treatment effects~\citep{stuart_matching_2010}. %(see Illustration~\ref{fig:intro_panel}). % Both issues increase extrapolation bias, Add to theoretical insights?. % MFP: TODO: define manifold
%\citep{goos_surprising_2001, abadie_using_2021, roberts_adjusting_2020}

% \paragraph{On the importance of geometry}

% \begin{itemize}
%     \item Geometric structure is an important bias in ML.
%     \item Structural causal models entail geometric structure, geometric structure reflects causal information. 
%     \item Geometry reflects complexity, functional relationship between different R.V.s.
%     \item In (Dominguez-Olmedo et.al, 2023), the authors show the benefits of accounting for the geometry in the generation of causally-grounded counterfactual explanations for ML classifiers.
%     \item Indeed, Riemannian metrics reflect causal knowledge embedded in a SCM~\cite{dominguez-olmedo_data_2023}.
%     \item \cite{dominguez-olmedo_data_2023} shows that structural causal models (SCM) induce geometric structure Under certain mild conditions, causal structure translates into geometric information captured by data manifolds. They derive sufficient conditions, Notable classes of/common classes of SCMs satisfy such conditions.
%     \item Given a Riemannian metric, we can define distance-based operations (like matching for TE estimation) that are informed by the causal structure.
% \end{itemize}

%\begin{wrapfigure}{r}{0.5\textwidth}
\begin{figure}[ht]
  \vspace{-0.3cm}
  \begin{center}
    \includegraphics[width=\textwidth]{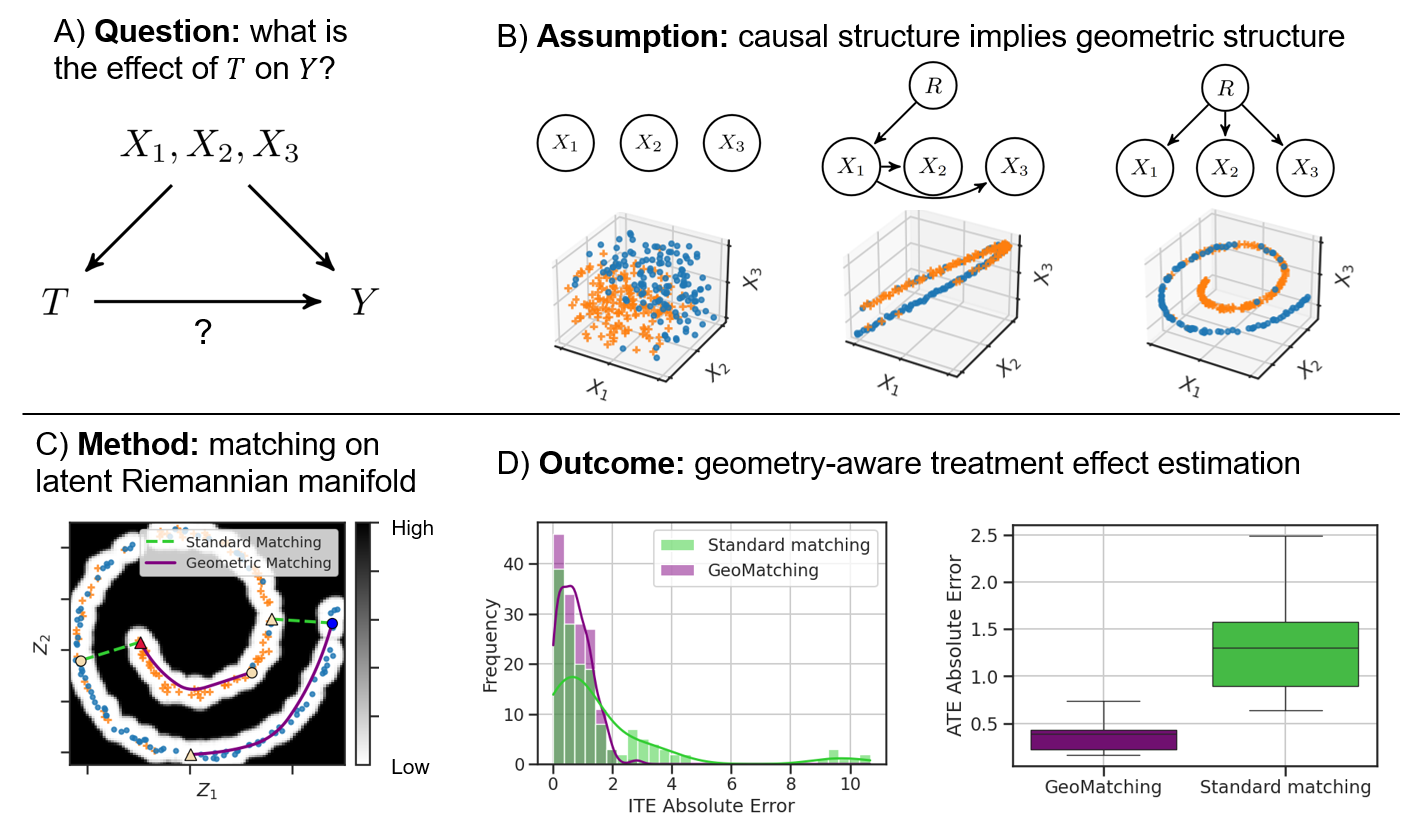}
  \end{center}
  \vspace{-0.3cm}
  \caption{\textbf{Key idea of the paper}. We propose \textit{GeoMatching}, a geometry-aware matching framework that pairs cases and controls along the manifold of confounders $X$. A) Assumed causal graph in this work; B) Different sub-causal graphs for covariates $X$ often induce different geometric data structures; latent variable $R$ represents the manifold structure; C) GeoMatching learns a latent Riemannian metric $\G$ (represented as colorbar) which captures the geometric structure and uncertainty of the data, enabling matched pairs to be faithful to the manifold structure.
  %Comparison of Standard matching and GeoMatching (our contribution) in terms of matched units and treatment effect estimation performance.
  D) Geometry-aware matched samples translate into more accurate estimates of individual (ITE) and average treatment effects (ATE).
  %, according to average treatment effect (ATE) absolute error averaged across 5 random seeds. %\textbf{Step 1}: train a latent variable model to learn a latent representation $Z$ and Riemannian metric $\mathbb{E}[\mathbf{G}]$. \textbf{Step 2:} estimate treatment effects by matching units according to geodesic distances along the data manifold.
  \label{fig:intro_panel}}
  \vspace{-0.3cm}
%\end{wrapfigure}
\end{figure}
%\vspace{-0.3cm}

In this work, 
%we assume that observations are the manifestation of a much smaller set of hidden variables (disease, genetics, environmental factors, etc). For example, a particular disease will influence the content of several medical notes and lab tests, but ultimately, it is the absence/presence of the disease that matters.
we assume the confounders lie on a manifold of much smaller dimensionality, this is the so called \textit{manifold hypothesis}. Such manifold structure arises from underlying causal mechanisms that induce statistical correlations among covariates in a constrained manner~\citep{dominguez-olmedo_data_2023}, as illustrated in Figure~\ref{fig:intro_panel}B.
For example, a sport trainer might want to estimate the effect of a particular training strategy (treatment) on athletes performance (outcome). To estimate such effect using observational data, the trainer might need to adjust for a wide variety of covariates, including height, weight and physiological measurements (confounders). Here, two athletes should be more or less similar according to natural differences in their physiology's: ideally, distances should respect the manifold structure
%-- which is generally not the case for Euclidean distance --
-- here, anatomical constraints -- such that confounding bias in treatment effect estimation can be reduced.  
% TODO: refine example, shorten.
%
%\item \textbf{Objective}: learn geometry-aware latent representation where distances between observations are meaningful, enabling downstream applications such as estimation of treatment effects.
%
%We exploit geometry to capture the intrinsic underlying data structure of the confounders distribution, so that confounding bias can be reduced.
%

We propose \textit{GeoMatching}, a framework to estimate treatment effects accounting for the underlying geometry and uncertainty of the data. Instead of working in the space of raw input characteristics, we learn a low-dim latent representation together with a latent \textit{Riemannian metric} (a generalization of the Euclidean metric) such that raw distances along the original manifold embedded in the high-dim input space are preserved in the low-dim latent space (see Figure~\ref{fig:intro_panel}C). We can then leverage those \emph{geometry-aware} distances for matching to estimate treatment effects more accurately, as shown in Figure~\ref{fig:intro_panel}D.\footnote{Figure~\ref{fig:intro_panel}D shows ITE/ATE estimation results corresponding to the synthetic 3D-swissroll dataset, which are later expanded in Section~\ref{subsec:swissroll}. Appendix~\ref{app:datasets} describes the data generation process in detail.}
The latent representation is a \textit{Riemannian manifold}, where distance between two data points is given by minimum path length along the data manifold. To our knowledge, this work is the first one looking at geometry-aware methods (in a differential geometric sense) for causal inference.

%by performing matching in a latent space according to a Riemannian metric (instead of Euclidean) that preserves the geometry of the data manifold in the original input space.
%
%
%We resort to a Gaussian Process Latent Variable model (GPLVM) to learn a latent representation of input covariates while accounting for the uncertainty in the dimensionality reduction step, which is then accounted for in the learned latent Riemannian metric.
%
% We could use a VAE as well~\cite{arvanitidis_latent_2021}, GPs have two advantages: i) suitable inductive biases for uncertainty estimation as we go away from data manifold, ii) closed-form prior distribution for tensor metric thanks to GPs being closed under linear operators (e.g., derivatives).
%
%Figure~\ref{fig:intro_panel} gives an overview and intuition of the proposed approach. The first row depicts the key underlying assumption of our work, i.e., that causal structure implies geometric data structure.
%The second row illustrates the benefit of using a Riemannian metric (instead of Euclidean) to account for data uncertainty: matched pairs will be faithful to the manifold structure (a), which translates into more accurate (less biased) individual and average treatment effect estimators (d-e).

\paragraph{Contributions} The contributions of this work are the following: first, we introduce \textit{GeoMatching}, a geometry-aware matching method that captures the intrinsic data structure of the confounders leveraging Riemannian geometry, accounting for data uncertainty with particular robustness to outliers. We describe assumptions under which we expect GeoMatching to improve over standard matching methods. Second, we empirically show that GeoMatching yields better results compared to other baselines on diverse synthetic and real-world datasets. We show its effectiveness as we increase the input space dimensionality, in the presence of outliers, or in semi-supervised scenarios.

\section{Standard Matching  and Problem Description}\label{sec:background}

\paragraph{Notation.}
Let $(X,T,Y)$ denote input (pre-treatment) covariates $X \in \mathcal{X}$, treatment variable $T \in \{0,1\}$ and outcome variable $Y \in \mathcal{Y}$. We operate in the standard potential outcome framework introduced by~\cite{rubin_causal_2005}. That is, we assume that each observation has two potential outcomes $Y(0)$ and $Y(1)$ of which only one is observed, $Y = (1-T)Y(0) + TY(1)$, which is the outcome corresponding to the administered treatment $T$.
Concretely, imagine that a single individual with pre-treatment covariates $X_\star \in \mathbb{R}^d$ is exposed to a treatment $T_\star = 1$, and we observe their outcome $Y_\star(1) \in \mathbb{R}$; this is called the \textit{treated unit}. We also assume access to covariates $X_j$ and outcomes $Y_j(0) \in \mathbb{R}$ of $m$ individuals not exposed to treatment ($T_j = 0$) -- the pool of available control units.
Let $\mu_t(x)=\mathbb{E}[Y(t) \given X=x]$ denote the expected outcome under treatment $T=t$.
%Let $\mu_1(x)=\mathbb{E}[Y(1) \given X=x]$ and $\mu_0(x)=\mathbb{E}[Y(0) \given X=x]$ denote the expected outcome with or without treatment.
%
We make the following standard assumptions from causal inference~\citep{pearl_causal_2009}:
%
%i) \textit{ignorability}: $Y_i(1), Y_i(0) \perp T_i \given X_i$, which means that there are no hidden confounders, ii) \textit{consistency}: $\forall t, T_i=t \rightarrow Y_i(t) = Y_i$, which ensures that $Y_i(t)$ is the observed outcome when $T_i=t$, and iii) \textit{overlap}: $\forall x, 0 < P(T_i=1 \given X_i = x) < 1$, which states that there is a chance of receiving either the treatment or control for any unit.
\begin{enumerate}[label=\roman*), noitemsep, topsep=0pt]
    \item \textit{ignorability}: $Y_i(1), Y_i(0) \perp T_i \given X_i$ -- there are no hidden confounders.
    \item \textit{consistency}: $\forall t, T_i=t \rightarrow Y_i(t) = Y_i$ -- $Y_i(t)$ is the observed outcome when $T_i=t$.
    \item \textit{overlap}: $\forall x, 0 < P(T_i=1 \given X_i = x) < 1$ -- there is a chance of receiving either the treatment or control for every unit.
\end{enumerate}
% Strong ignorability is a sufficient condition for causal identifcation (Rosenbaum and Rubin 1983; Imbens and Wooldridge 2009).

\paragraph{Problem Formulation.} 
The main goal of matching is to infer the effect of the treatment by comparing the treated outcome $Y_\star(1)$ to that of a \textit{matched unit} $\hat{Y}^{\text{match}}_\star$ that plays the role of a hypothetical ``twin'' of the treated unit had they not been exposed to treatment. In other words, $\hat{Y}^{\text{match}}_\star$ aims to be as close as possible to their (unobserved) \textit{potential} outcome $Y_\star(0)$. More formally, we want the matched unit to approximate the expected outcome without treatment, 
$\mu_0(x)=\mathbb{E}[Y(0) \given X=x]$, 
i.e., $\mathbb{E}[\hat{Y}^{\text{match}}_\star] \approx \mu_0(X_\star)$.
A matched unit is constructed as an average of the outcomes of a few control units. 
In the following and without loss of generality, we focus on the nearest neighbor matching, i.e., the case when a single control unit gets assigned to the treated unit of interest. Let $j^{\dagger}$ denote the index of the nearest neighbor (hypothetical ``twin''). %Equation~\eqref{eq:weigth_matching} can then be written as
A matched unit can then be built as:
\begin{equation}
    \hat{Y}^{\text{match}}_\star = \sum_{j:T_j=0} w_{j} Y_j, \quad\text{where}\quad \begin{cases}
w_j = 1 & \text{if } j=j^{\dagger} \text{ (matched)} \\
0 & \text{if } j \neq j^{\dagger} \text{ otherwise}.
\end{cases}\label{eq:Y_match}
\end{equation}
Note that, if we allow $w_j$ to take continuous values, i.e., $0 \leq w_j \leq 1$, and $\sum_j w_j = 1$, we recover the synthetic controls framework~\citep{abadie_using_2021, curth_cautionary_2024}.

% Estimating treatment effects from observational data suffers from the issue of \textit{confounding bias} which occurs when the distribution of covariates for cases and controls is different. To address such problem, standard matching methods aim to ... COMPLETE
%
% The goal of a matching procedure is to assign one or more control units $j$ to a treated unit $i$ (and viceversa) based on the similarity between their covariates $X_i$ and $X_j$. 
%
To find the best covariate match $X_{j^\dagger}$ for treatment unit $X_{\star}$ we perform nearest neighbours search among all control units according to some distance metric $d(\cdot,\cdot)$, namely:%~\citep{stuart_matching_2010}
\begin{eqnarray}
    %\underset{j \in [m]}{\text{argmin}} \; d_{\mathcal{E}}(X_{\star}, X_j) =
    j^{\dagger} = \underset{j \in [m]}{\text{argmin}} \; d(X_{\star}, X_j).\label{eq:problem_formulation}
\end{eqnarray}
The most popular choice is the squared Euclidean distance, i.e., $d(X_{\star}, X_j) = ||X_{\star} - X_j ||^2$, which we refer to as \textit{standard matching method}. However, if there is underlying structure in the confounders given by the natural causal mechanisms of the problem,\textit{ which is the best distance to use and under which assumptions?}
%But is this the best distance? If there is underlying structure in the confounders, which distance should we rely on?
%
This question motivates our contribution GeoMatching, which we describe next.
%
%To answer, let us dig into some key concepts of Riemannian geometry.

% Standard matching can be performed with or without replacement, and one may enforce a caliper, i.e. a maximal distance between matches.
%
%citing Clivio et.al: "Alternatively, one might consider all matches simultaneously through an optimisation programme (optimal match) (Rosenbaum, 1989)."
% %

\section{GeoMatching}\label{sec:geomatching}

% \subsection{Key Riemannian Concepts}\label{sec:background_riemannian}

% As stated in Equation~\ref{eq:problem_formulation}, matching requires computing the distance $d(X_\star, X_j)$ between treated unit $X_\star$ and control unit $X_j$, illustrated in Figure~\ref{complete}.

% We can define a distance as a length of a curve. For example, Euclidean distance corresponds to the length of a straight line. But what if data lies on a potentially curved space, i.e., a non-linear manifold?

% A Riemannian metric $\G$ encapsulates an infinitesimal notion of distance on a manifold $\mathcal{M}$. More formally, 
% $\G$ defines at each point $x \in \mathcal{M}$ a smoothly varying inner product in the tangent space $T_x\mathcal{M}$. The inner product is defined as:
% \begin{equation}
%       \langle x_1,x_2 \rangle_x = x_1^{\mathrm{T}}\G(x)x_2,
% \end{equation}
% where $x_1, x_2 \in T_x\mathcal{M}$ and $x \in \mathcal{M}$.
% % TODO: Add examples for Euclidean and Mahalanobis.

% The length of a smooth curve $\gamma: [0,1] \rightarrow \mathcal{M}$ is then defined as:
% \begin{equation}
%     \mathrm{Length}(\gamma) = \int_0^1 \sqrt{\gamma'(t)^T \G(\gamma(t))\gamma'(t) dt},
% \end{equation}
% where $\gamma'(t) = \frac{d}{dt}\gamma(t)$ denotes the velocity of the curve. The \textit{Riemannian distance} between two points on manifold $\mathcal{M}$ is defined as the length of the geodesic curve, which is the  length-minimizing curve connecting them.

\subsection{Assumptions}\label{sec:motivation}

%\subsection{Assumptions}\label{sec:conditions}
%GeoMatching relies on two key underlying assumptions.

\paragraph{Manifold Hypothesis}
%In this work, we assume the \textit{manifold hypothesis}, which states that the high-dimensional confounders $X$ lie near a low-dimensional (non-linear) manifold embedded in the original input space~\citep{whiteley_statistical_2024}.
%
The \textit{manifold hypothesis} states that the high-dim confounders $X$ lie near a low-dim (non-linear) manifold embedded in the original input space~\citep{whiteley_statistical_2024}.
Low-dimensional structure arises due to constraints from physical laws and nature, or in other words, geometry often results from the underlying causal mechanisms between different covariates~\citep{dominguez-olmedo_data_2023}, as illustrated in Figure~\ref{fig:intro_panel}B.
%
% Under certain mild conditions, causal structure translates into geometric information captured by data manifolds. Notable classes of SCMs satisfy such conditions, as shown by Dominguez.
%
The manifold hypothesis is satisfied under common classes of Structural Causal Models; data types where it is often fulfilled include images of 3D objects, 
%under different angles and light conditions,
phonemes in speech signals, or video streams~\citep{fefferman_testing_2016}.
% The \textit{manifold hypothesis}, states that most naturally occurring datasets lie near a non-linear manifold embedded in the
%original input space~\citep{fefferman_testing_2016, dominguez-olmedo_data_2023}.

\paragraph{Geometric Faithfulness}
%In the case where the treatment effect varies smoothly along the manifold of the confounders, a natural choice is to perform matching leveraging distances along the manifold structure. 
We assume that the treatment effect varies continuously along the manifold of confounders.
Continuity ensures that points close to each other in the latent manifold will have similar treatment effects. This property is central in GeoMatching, as it allows us to ensure that small changes in the input (manifold points) do not lead to large, abrupt changes in the treatment effect.
%
%guarantee that when the treatment effects are computed over the manifold instead of the original coordinate system of the original problem, information about the treatment effect is preserved under the new coordinate system.
%
Back to the motivational example in Section~\ref{sec:intro}, we expect athletes with similar anatomical constraints (reflected by the manifold structure) to respond similarly to different training strategies. As another example, a chemist might want to assess the effect of adding a catalyst to a chemical reaction (treatment) on the chemical reaction rate (outcome) given experimental conditions such as pH, purity of reactants, pressure, temperature, etc (confounders). As we move along the space of possible experimental conditions, we expect a smooth variation of the treatment effect.

%In Appendix~\section{}

\subsection{A Geometric Take on Treatment Effect Matching}
%\paragraph{Key Riemannian Concepts.}
The distance between a treated and control unit can be defined as the length of a curve.
%, as illustrated in Figure~\ref{fig:cartoon_distance}.
A Euclidean distance for example corresponds to the length of a straight line. In general, if confounders $X$ lie on a manifold $\mathcal{M}$ (potentially curved space), we define the length of a smooth curve $\gamma: [0,1] \rightarrow \mathcal{M}$ as:
\begin{equation}
    \mathrm{Length}(\gamma \,; \G) = \int_0^1 \sqrt{\gamma'(r)^T \G(\gamma(r))\gamma'(r) dr},
\end{equation}
where $\gamma'(r) = \frac{d}{dr}\gamma(r)$ denotes the velocity of the curve, and $\G$ denotes a \textit{Riemannian metric}, which is a smooth function that assigns a symmetric positive definite matrix $\G(x)$ to different locations $x \in \mathcal{M}$.
Intuitively, $\G$ encapsulates an infinitesimal notion of distance on manifold $\mathcal{M}$.

Standard matching relies on a Riemannian metric $\G = I$, resulting in the squared Euclidean distance:
\begin{equation}
    d_{\mathcal{E}}(X_\star, X_j) = %||X_\star - X_j||^2 = 
    (X_\star - X_j)^T \mathbf{I} (X_\star - X_j).
\end{equation}
Mahalanobis matching methods rely on a constant Riemannian metric $\G = \Sigma$,
%i.e., $d_{\text{Mahalanobis}}(X_\star, X_j) = (X_\star - X_j)^T \boldsymbol{\Sigma} (X_\star - X_j)$.
\begin{equation}
    d_{\mathcal{M}}(X_\star, X_j) = (X_\star - X_j)^T \boldsymbol{\Sigma}^{-1} (X_\star - X_j).
\end{equation}

GeoMatching relies on a \textit{Riemannian distance} $d_{\mathcal{R}}(X_\star, X_j)$, a generalization of the Euclidean and Mahalanobis distance that replaces the constant covariance matrix $\mathbf{I}$ or $\boldsymbol{\Sigma}$ by an input-dependent covariance matrix $\G(x)$, and integrates out the infinitesimal distance contributions along the shortest curve -- also called \textit{geodesic curve} -- connecting the two confounders along the manifold.
%length-minimizing curve connecting the two confounders:
% \begin{equation}
%  d_{\text{Riemannnian}}(X_{\star}, X_j)  =  \underset{\gamma_j}{\text{min}} \; \mathrm{Length} \Big( \gamma_j \,; \mathbf{G} \Big).
%  \label{eq:geodesic_distance}
% \end{equation}
\begin{equation}
 d_{\mathcal{R}}(X_{\star}, X_j)  =  \underset{\gamma_j}{\text{min}} \; \mathrm{Length} \Big( \gamma_j \,; \mathbf{G} \Big).
 \label{eq:geodesic_distance}
\end{equation}
%
%The \textit{Riemannian distance} between two points on manifold $\mathcal{M}$ is defined as the length of the \textit{geodesic curve}, which is the length-minimizing curve connecting those two points along the manifold.
See Appendix~\ref{app:riemannian} for further details and precise mathematical definitions of key Riemannian concepts.

\subsection{Algorithm Description}\label{sec:livm}

%Under the manifold hypothesis and manifold causal coherence stated in Section~\ref{sec:conditions}, and in addition to the standard causal assumptions of non-hidden confounders, consistency, ignorability and overlap), choosing the Riemannian distance yields better estimation of treatment effects compared to any other distance metric, including the Euclidean distance at the core of the standard matching method.
We can now introduce GeoMatching, a geometry-aware matching framework that leverages Riemannian distances to pair treated and control units.
% , summarized in Algorithm~\ref{alg:geomatching}.

%\item \textbf{Step 1}: Learn Riemannian metric $\mathbf{G}$ in a latent space $\mathcal{Z}$ given confounders $X \in \mathcal{X}$.
\paragraph{Step 1: Learn latent Riemannian manifold.} GeoMatching learns a topology-preserving low-dimensional representation $Z \in \mathcal{Z}$ and latent Riemannian metric $\mathbf{G}$ given confounders $X \in \mathcal{X}$. Considering a latent representation instead of the original input space makes it easier for metric learning, specially if the input space is high-dimensional, and enables to filter out irrelevant information unrelated to the geometry of the confounders. Several methods can be adopted for this stage. One option is to assume a  linear projection using Principal Component Analysis (PCA)\footnote{Note that PCA is not guaranteed to preserve geometric structure in general, for which other algorithms such as Isomap or Multidimensional Scaling might be more suitable.} and fit a parameterized Riemannian metric $\mathbf{G}$.
An alternative is to train a probabilistic latent variable model to jointly learn $Z$ and $\mathbb{E}[\mathbf{G}]$, propagating Jacobian information~\citep{tosi_metrics_2014}. % by propagating Jacobian information, via the pull-back metric.
In the experimental Section~\ref{sec:experiments}, we resort to the former strategy for simplicity, and posit a parametric Local Inverse Variance (LIV) Riemannian metric~\citep{arvanitidis_locally_2016} in the latent space $\mathcal{Z}$, which is defined as the inverse of a local diagonal covariance matrix with $j$-th entry defined as:
\begin{equation}
    \G_{jj}(z) = \left(\sum_{i=1}^N w_i(z) \left(z_{ij} - z_j \right)^2 + \rho \right)^{-1}, \quad \text{where} \quad w_i(z) = \mathrm{exp} \left( -\frac{||z_i - z||^2}{2 \sigma^2} \right).
    \label{eq:livm}
\end{equation}
Parameter $\sigma$ controls how fast uncertainty increases as we move away from the manifold; parameter $\rho$ is needed for numerical instability, influencing the magnitude values that the metric can take. 
LIV assumes a Euclidean metric locally, which is then regularized to have large volume measure in regions of the feature space far away from observations. The inductive bias underlying such metric is that shortest paths on the data manifold should remain close to the observed data, we want curves to avoid crossing low-density high-uncertainty regions. A key advantage of this metric is that geodesic computation is less costly, as off-diagonal terms in metric $\G$ are zero.

% Arvanitidis 2022: "The non-parametric space is constructed using a local metric that is the inverse of a local covariance matrix. Here locality is defined via a Gaussian kernel, such that the manifold learning can be seen as a form of kernel smoothing. This indicates that our scheme for learning a manifold might not scale to high-dimensional input spaces".

% Interpretation of locally Euclidean metric in Z space: the intuition is that similarity of observations should be measured based on similarity of their latent variables, in a locally Euclidean sense.

% LVI metric is a deterministic manifold, we also consider a stochastic/statistical manifold...

\paragraph{Step 2: Compute Riemannian distances.} We compute geodesic or Riemannian distances in the latent representation between treated and control units:
\begin{equation}
 d_{\mathcal{R}}(X_{\star}, X_j)  =  \underset{\gamma_j}{\text{min}} \; \mathrm{Length} \Big( \gamma_j \,; \mathbf{G} \Big),
 \label{eq:geodesic_distance2}
\end{equation}
where $\gamma_j: [0,1] \rightarrow \mathcal{Z}$ is a curve connecting the latent representations of the confounders, %the unit of interest $X_\star$ with datapoint $X_j$,
i.e., $\gamma_j(0)=Z_{\star}, \, \gamma_j(1)=Z_{j}$.
To compute geodesic distances, we instantiate a parameterized (discretized) geodesic curve\footnote{Considering a discretized curve for the geodesic scales more gracefully than other graph-based approaches, as the discretization of the curve is always one-dimensional independently of the dimension of the latent space.} and optimize its parameters by minimizing the following second-order ordinary differential equation~\citep{do_carmo_riemannian_1992}:
\begin{equation}
    \gamma'' = -\frac{1}{2}\G^{-1} \left[ \frac{\partial \, \mathrm{vec} \, \G}{\partial \, \gamma} \right]^T (\gamma' \otimes \gamma'),
\end{equation}
where $\mathrm{vec}\, \G$ stacks the columns of $\G$ and $\otimes$ denotes the Kronecker product. Based on Picard-Lindelöf theorem~\citep{tenenbaum_ordinary_1985}, geodesics are guaranteed to exist and are locally unique given a starting point and initial velocity $\gamma'$. See Appendix~\ref{sec:computational} for further details and discussion on computational cost of GeoMatching.

\paragraph{Step 3: Matching along Riemannian manifold.} Given the Riemannian distances from Equation~\eqref{eq:geodesic_distance2}, we match treated and control units along the Riemannian manifold by minimizing:
\begin{equation}
    j^{\dagger} = \underset{j \in [m]}{\argmin} \; d_{\mathcal{R}}(X_{\star}, X_j).
    \label{eq:matching} 
\end{equation}

\paragraph{Step 4: Estimate causal treatment effects.} Given the matched units $\hat{Y}^{\text{match}}_\star$ defined in Equation~\eqref{eq:Y_match}, we can estimate the individual and average treatment effects as follows:
\begin{eqnarray}
    \widehat{\mathrm{ITE}} & = & Y_\star - \hat{Y}^{\text{match}}_\star, \\
    \widehat{\mathrm{ATE}} 
    %= \mathbb{E}_{X_\star \sim \mathcal{D}} \left[ \mathrm{ITE(X_\star)}  \right]
    & = & \frac{1}{N} \sum^N_{i=1} \left( Y_i - \hat{Y}^{\text{match}}_i \right)^{T_i=1} \left( \hat{Y}^{\text{match}}_i - Y_i\right)^{T_i=0}.\label{eq:te_estimation} 
\end{eqnarray}

\section{Related Work}\label{sec:related}

\paragraph{Matching in Latent Representations.}

Several previous works learn low-dim representations before matching to mitigate undesirable effects of high dimensions in causal inference.
\cite{clivio_neural_2022} leverage the representation of intermediate layers of a neural network that predicts the propensity score $p(T|X)$, showing that such representation is approximately a balancing score, i.e., $X \perp T | Z$.
\cite{luo_matching_2017} and \cite{zhao_matching_2022} propose matching methods on linear projections of the data based on the sufficient dimensionality reduction property, i.e., $X \perp Y | Z$.
%
%Also, latent representations only have info related to propensity score, risk of matching at random.
%
\cite{wang_flame_2021} and \cite{li_matching_2016} propose matching algorithms for high-dim data: the former relies on variable selection for categorical data; the later conducts and aggregates matching on random projections of the data.
%a matching algorithm for high-dimensional categorical data, scaling to huge datasets with millions of observations. They propose an algorithm for variable selection, learn a Hamming distance matrix that indicates which covariates should be included in distance computation when matching covariantes.
%
%\cite{li_matching_2016} provide a randomized matching method for high-dimensional data based on random projections of the data.%, the idea is to perform random projections of the data, conduct standard matching in each of these, and compute median counterfactual outcome.
    %They do not account for geometry, and only handle discrete covariates.
%
 \cite{clivio_towards_2023} characterize the bias added to a weighting estimator when using a data representation and propose an objective to learn suitable representations for causal inference.
In contrast to our work, none of these methods account for the geometry of confounders.%\footnote{We note that GeoMatching is a framework rather than a closed-form procedure, and improvements on learning better latent representations can be armored with Riemannian metrics, as long as the latent representation preserves geometric structure of the confounders.}

\paragraph{Geometry-aware Machine Learning Models.}

Accounting for data geometry has been the object of study of several works in the machine learning community.
\cite{tosi_metrics_2014} propose to learn a latent representation equipped with a Riemannian metric by assuming a Gaussian Process prior on the mapping function from latents to observations.
\cite{arvanitidis_latent_2021}, \cite{shao_riemannian_2017} and \cite{chen_metrics_2018} propose to exploit the geometry of the latent representation in deep generative models.
%
% TODO: add \cite{chadebec_geometry-aware_2020}: Riemannian metric for the latent space of VAEs.
%
%
%All conclude that treating the latent space as a Riemannian manifold instead of a Euclidean space entails several benefits, including better classification performance,  interpolation and density estimation.
These works conclude that treating the latent space as a curved Riemannian space instead of a Euclidean space yields several benefits for clustering, interpolation and prediction. 
%
%, including classification performance, interpolation, density estimation, or generation of random walks.
%
%Some works propose to learn a metric using a Latent Variable Model~\citep{tosi_metrics_2014, arvanitidis_latent_2021, chen_metrics_2018, shao_riemannian_2017}.
A caveat of those methods is that the computation of geodesics is expensive and selection of hyperparameters for downstream tasks is often non trivial.
%
%\item \cite{arvanitidis_latent_2021}: given an arbitrary generator, derives metric + theoretical analysis. Also fixes variance estimates away from data. Take-away: treating the latent space as a curved space instead of a Euclidean space yields several benefits for clustering, interpolation and prediction. Generalizes work from~\cite{tosi_metrics_2014} with GPLVMs to arbitrary generators, does not require learning. Geodesics computed in matlab by solving ODE.
%
More recently, \cite{dominguez-olmedo_data_2023} show the benefits of accounting for the geometry in the generation of causally-grounded counterfactual explanations for ML classifiers.
%\end{itemize}

%MFP: TODO add:
% Applications domains where geometric structure matters: computer vision, human motion capture, robotics and protein sequencing. Benefits of taking into account the geometry: kNN clustering, latent probability distributions, interpolations, random walks, contrastive learning.
% Detlefsen, N. S., Hauberg, S., and Boomsma, W. Learning meaningful representations of protein sequences. Nature communications, 13(1):1–12, 2022

%\paragraph{Accounting for geometry in causal inference.}
\paragraph{Geometry-aware Causal Inference.}
While there has been extensive work on learning suitable latent representations for causal inference, very few have considered the geometry of the latent space for treatment effect estimation. \cite{yan_exploiting_2024} is the only work we found that proposes to consider the geometry of confounders to improve TE estimation. They propose an optimal transport framework to promote the assignment of low weights for outliers, but they still assume Euclidean distances as opposed to a generic Riemannian distance along the data manifold. To our knowledge, our work is the first one to propose geometry-aware TE estimation based on Riemannian geometry.

\section{Experiments and Results}\label{sec:experiments}

We evaluate GeoMatching on synthetic and real-world datasets against other matching strategies in a  synthetic swissroll toy scenario, a motion capture data, and two standard causality benchmark datasets. We show that GeoMatching mantains stability as we increase input dimensionality or in the presence of outliers, and benefits from semi-supervised scenarios.%\footnote{We provide open-source code and scripts to reproduce results at: \texttt{<Anonymized>}.}
%
% We consider a synthetic swissroll toy scenario, motion capture data, and three standard causality benchmark datasets. In all cases, we demonstrate improvements in treatment effect estimation. We show that GeoMatching mantains stability as we increase input dimensionality or in the presence of outliers, and benefits from semi-supervised scenarios, in contrast to other approaches.
% %

\subsection{Experimental Setup}\label{subsec:experimental-setup}

%GeoMatching consists of two main stages: first, we learn a low-dimensional latent representation equipped by a Riemannian metric and second, we perform matching in that latent space according to Riemannian distances, as described in Section~\ref{sec:geomatching}. For the first stage,
In order to learn the latent Riemannian manifold (Step 1 in Section~\ref{sec:geomatching}), 
%we opt for the first strategy presented in Subsection~\ref{sec:livm}, namely,
%
we project the data using a deterministic dimensionality reduction technique (either PCA or Isomap)
% 
% we learn a non-parametric metric space by projecting the data using a deterministic dimensionality reduction technique (either PCA or Isomap) 
%
and posit 
%a Local Inverse Variance (LIV) Riemannian metric (see Equation~\ref{eq:livm}),
a parametric smoothly changing kernel-based Riemannian metric defined by Equation~\eqref{eq:livm}. This choice is motivated twofold: i) the resulting pipeline is very simple and easy to implement, with effectively a single parameter $\sigma$ for model selection, ii) we also explored learning the latent Riemannian metric using a Gaussian Process latent variable model, but found that it was less suitable in practice in terms of computational cost and quality of results. %, see Appendix~\ref{app:gplvm} for further results with a GPLVM.
In terms of compute resources, we used 1 CPU for each run, with 10 GB and running time of < 24h for each experiment.

\paragraph{Baselines.} We evaluate the performance of GeoMatching according to two dimensions: i) which space gets considered for matching (the original input space $\mathcal{X}$, or a latent space $\mathcal{Z}$), and ii) which distance we rely on (either Euclidean, Mahalanobis, or Riemannian). We also include random matching to have a reference baseline on how difficult it is to match treated and control units on that particular dataset.
%We compare our approach, denoted \texttt{Geomatching (Z)}, against the following baselines: Random matching (\texttt{Random}), standard matching in $\mathcal{X}$ (\texttt{Euclidean in X}), standard matching in latent space $\mathcal{Z}$ (\texttt{Euclidean in Z}), and an ablation of GeoMatching applied to the original space directly without any prior dimensionality reduction, denoted as \texttt{Geomatching (X)}.
In terms of latent representation $Z \in \mathcal{Z}$, we consider three alternatives: PCA, Isomap, and the latent representation of Neural Score Matching (NSM) from~\cite{clivio_neural_2022} which corresponds to the first layer of a neural network that predicts treatment assignment. Accounting for all combinations of matching space and considered distance, this gives us a total of 9 baselines and 3 ablations of GeoMatching.

\paragraph{Evaluation.} For evaluation, we qualitatively inspect the matched units (geodesic curves) in the latent space, and quantitatively assess performance of treatment effect estimation via Average Treatment Effect (ATE) absolute error, defined as the absolute difference between the true and estimated ATEs. We also report Precision in Estimation of Heterogeneous Effect (PEHE)  and distributions of Individual Treatment Effect (ITE) in Appendix~\ref{app:empirical}.
When plotting the latent representation of GeoMatching, we visualize the \textit{magnification factor} -- also called Riemannian volume -- which measures the magnitude of the local distortion of space at location $p \in \mathcal{M}$, i.e., $V_{\G}(p) := \sqrt{\mathrm{det}\, \G(p)}$.
Curves that traverse regions with high magnification factor tend to have larger lengths; it can also be seen as thr energy needed to traverse such space. Geodesics or shortest paths tend to avoid regions with high values of magnification factor, which reflect high uncertainty away from the observed data manifold.

\paragraph{Model Selection.} We use single nearest neighbour matching strategy with replacement for all considered matching methods. We split our data 75/25 into train and test set: we use the train set to optimize parameter $\sigma$ of the LIV Riemannian metric (which controls how fast the metric increases as we move away from the data manifold), by minimizing ATE absolute error in the train set.
For the real-world datasets, we also select the dimensionality $K$ of the latent representation based on performance on the train set, sweeping over $K=\{2,3,4,5,6\}$ for Lalonde dataset, and $K=\{2,4,6,10\}$ for IHDP. For synthetic datasets, we fix the latent dimensionality $K=2$.
we report performance in both Train and Test set, averaged over 20 training random seeds for synthetic/semi-synthetic datasets, and 5 different training random seeds for real-world scenarios.

 % Our code and experiments are open source and can be found online.
%\vspace{-0.1cm}
\subsection{Synthetic Swissroll Dataset}\label{subsec:swissroll}
%\vspace{-0.2cm}

We adapt the classic swissroll dataset (see Appendix~\ref{app:datasets} for precise details on the data generation process) for the task of treatment effect estimation. We generate $N=200$ observations where (pre-treatment) covariates $X$ follow a 3D-swissroll manifold, treatment assignment variables $T$ are Bernoulli draws from a smoothly varying sigmoid function along the swissroll manifold, and outcome variables $Y$ are drawn from linear models.
To assess the impact of input dimensionality on the different matching strategies, we embed the 3D-swissroll in a space of higher dimensionality $D \in \{3, 5, 10, 15, 25, 50, 100\}$, by adding additional noisy (irrelevant) input dimensions.

\begin{figure}[h!]
  \centering
  \begin{minipage}[b]{0.32\textwidth}
    \includegraphics[width=\textwidth]{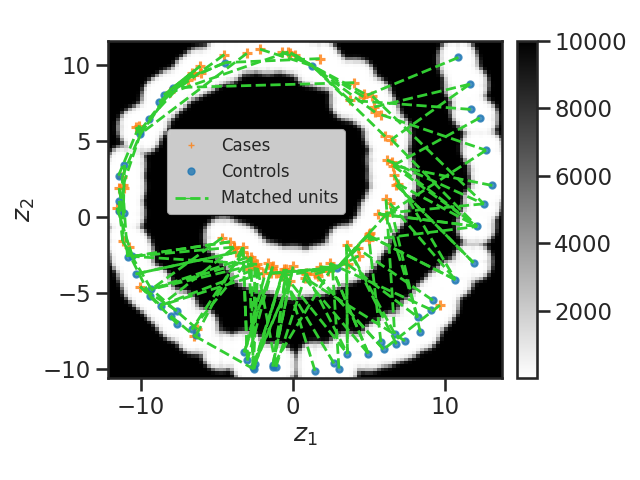}
    \caption*{(a) Euclidean in $\mathcal{X}$}
    \label{fig:swissroll_geo1}
  \end{minipage}
  \begin{minipage}[b]{0.32\textwidth}
    \includegraphics[width=\textwidth]{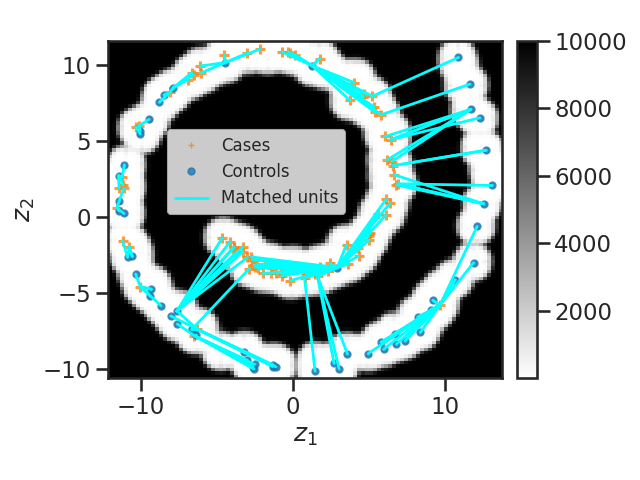}
    \caption*{(b) Euclidean in $\mathcal{Z}$}
    \label{fig:swissroll_geo2}
  \end{minipage}
  \begin{minipage}[b]{0.32\textwidth}
    \includegraphics[width=\textwidth]{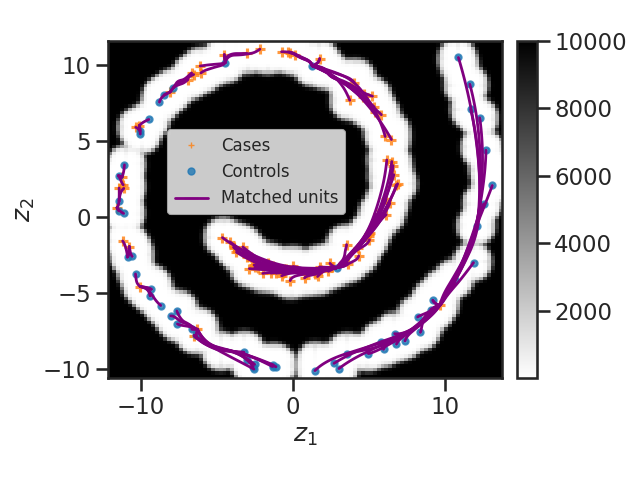}
    \caption*{(c) GeoMatching $\mathcal{Z}$}
    \label{fig:swissroll_geo3}
  \end{minipage}
  \hfill
  \vspace{0.3cm}
  \begin{minipage}[b]{0.43\textwidth}
    \includegraphics[width=\textwidth]{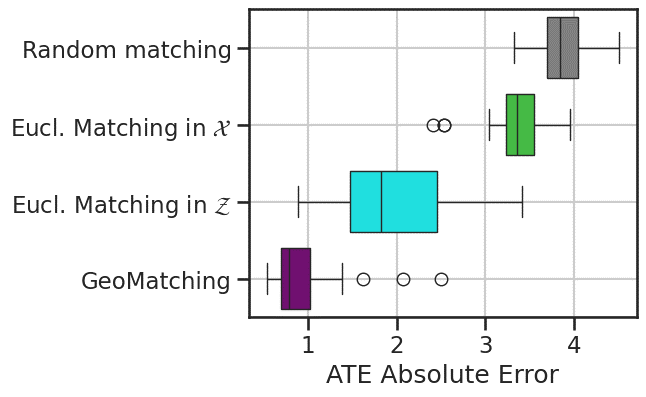}
    \caption*{(d) TE Estimation Performance}
    \label{fig:swissroll_ate_only25}
  \end{minipage}
  %\hspace{0.1cm}
  \begin{minipage}[b]{0.51\textwidth}
    \includegraphics[width=\textwidth]{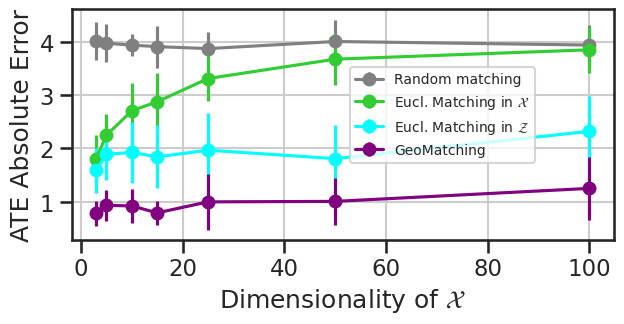}
    \caption*{(e) Impact of input space dimensionality $D$}
    \label{fig:swissroll_ate_highdim}
  \end{minipage}
  \caption{\textbf{Results for swissroll synthetic toy dataset}. (a-c) Matched pairs according to baselines and our method GeoMatching, (d-e) ATE Absolute Errors averaged over 20 random seeds.}
  \label{fig:swissroll}
  %\vspace{-0.4cm}
\end{figure}

For this dataset, we project the input covariates into a 2D latent space. Both PCA and Isomap return similar latent representations, preserving the geometry of the swissroll. Here we report results for PCA.
Figure~\ref{fig:swissroll} summarizes results on the swissroll dataset.  When inspecting matched units (geodesic curves) in the latent space (Figures~\ref{fig:swissroll}(a) to \ref{fig:swissroll}(c)), all baselines disregard the manifold geometry, pairing cases and controls across areas of high uncertainty. Instead, GeoMatching successfully couples cases and controls along the swissroll manifold by accounting for the confounders geometric structure and data uncertainty as we move away from the data manifold.

Looking at performance for TE estimation in Figures~\ref{fig:swissroll}(d) and~\ref{fig:swissroll}(e), GeoMatching yields the most accurate ATEs. As we increase the dimensionality $D$ of input space $\mathcal{X}$, performance of euclidean matching in $\mathcal{X}$ degrades until reaching the same performance as random matching. Unsurprisingly, euclidean matching in $\mathcal{Z}$ remains competitive as $D$ increases since it removes irrelevant dimensions (like GeoMatching) when projecting the data to a low-dim latent space.

%\vspace{-0.1cm}
\subsection{Semi-synthetic Human Motion Capture}
%\vspace{-0.2cm}

We use a semi-synthetic scenario of human motion capture data, which consists of real-world input covariates and synthetic treatment and outcome assignment variables.
Such setup lies in-between an ideal/simplified synthetic scenario (like the 3D-swissroll example from previous section) and a fully realistic scenario where there is no access to ground truth ITEs (like the LaLonde dataset in Section~\ref{sec:real-world-exp}).
The goal of this experiment is to demonstrate i) effectiveness of GeoMatching handling high-dim realistic covariates, and ii) robustness of GeoMatching against outlier confounders, in contrast to geometry-agnostic methods.

Human motion data has often been used to illustrate the impact of data geometry on downstream applications, as latent spaces are expected to be cylindrical or toroidal (doughnut-shaped)~\citep{tosi_metrics_2014, urtasun_topologically-constrained_2008}. 
Specifically, we consider motion 16 from subject 22 from the CMU Motion Capture Database\footnote{\url{http://mocap.cs.cmu.edu}} which is a repetitive jumping jack motion. Each observation corresponds to a human pose as acquired by a marker-based motion capture system.

We adapt the CMU mocap data for the downstream task of TE estimation. To motivate this new setup, consider \textit{Double Dutch}, a jump rope game that started in the streets and has now advanced to competitions, since 1974.\footnote{\url{https://youtu.be/iiEzf3J4iFk?feature=shared&t=49}}.
%so that they cross rhythmically. Two people rhythmically turn the ropes while the jumper(s) hop over them.\footnote{Double Dutch started as a street game, popularized by black women in NYC/ Black American experience, in communities with limited resources, but has now advanced to competitions since 1974 with awards and prizes.}
To play, two people turn two long ropes in opposite directions while, multiple players jump simultaneously, entering the rope area one at a time. Let $X$ be the motion capture of the current player jumping, $T$ the strategy of entrance, either slow ($T=0$) or fast ($T=1$), and $Y$ the probability of a successful entrance, when ropes do not trip up the new jumper's feet.
To make the data more challenging, we intentionally perturb some time frames to simulate outlier confounders in the dataset (i.e., players jumping in a weird, unnatural manner, or motion capture sensors being defective).
%
%We introduce outliers by artificially adding noise in a few rows of the covariate matrix $\mathcal{X}$.
%
See Appendix~\ref{app:datasets} for further details about the data generation process.
% A successful jump requires coordination between turners and jumpers.

\vspace{-0.1cm}
\begin{figure}[h!]
    \centering
    \begin{minipage}[b]{\textwidth}
    \includegraphics[width=\textwidth]{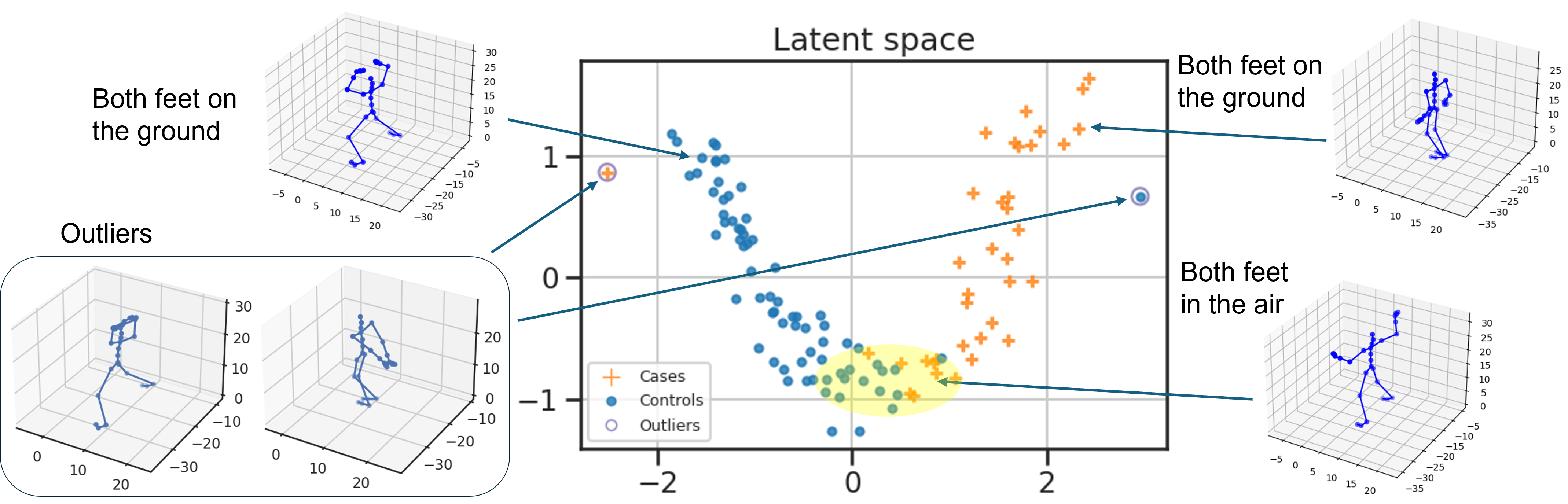}
    \caption*{(a) Data Description: CMU Motion Capture Data.}
    \label{fig:mocal_data}    
    \end{minipage}
    % \begin{minipage}[b]{0.24\textwidth}
    % %\includegraphics[width=\textwidth]{figures/mocap/placeholder.png}
    % \includegraphics[width=\textwidth]{figures/mocap/euclidean_X.png}
    % \caption{Euclidean in $\mathcal{X}$}
    % \label{fig:subfig1}
    % \end{minipage}
  \begin{minipage}[b]{0.28\textwidth}
    \includegraphics[width=\textwidth]{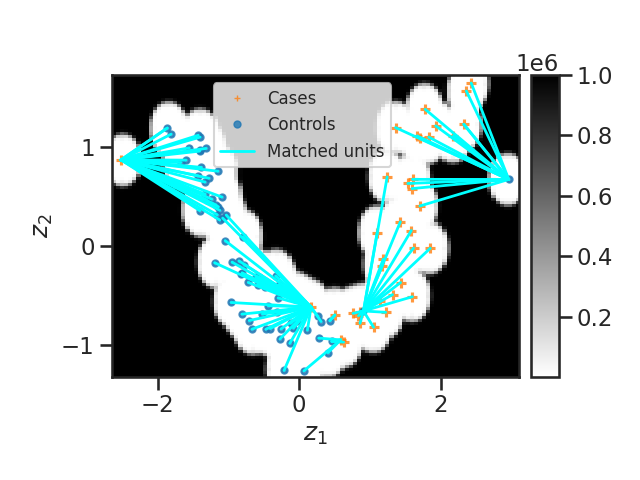}
    \caption*{(b) Euclidean in $\mathcal{Z}$}
    \label{fig:mocap_geo1}
  \end{minipage}
  \begin{minipage}[b]{0.28\textwidth}
    \includegraphics[width=\textwidth]{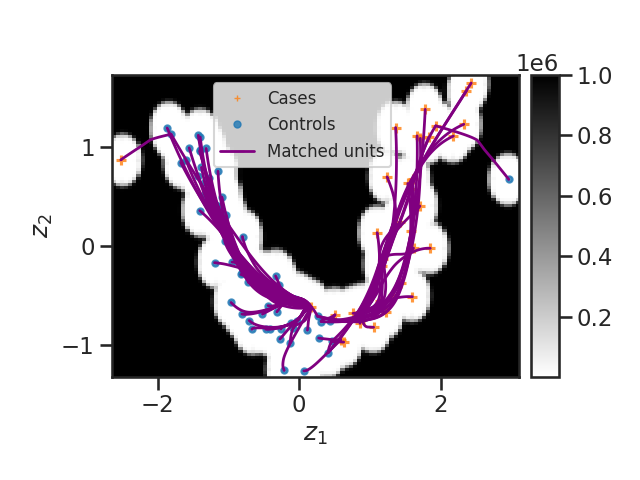}
    \caption*{(c) GeoMatching in $\mathcal{Z}$}
    \label{fig:mocap_geo2}
  \end{minipage}
  \begin{minipage}[b]{0.39\textwidth}
    \vspace{0.1cm}
    \includegraphics[width=\textwidth]{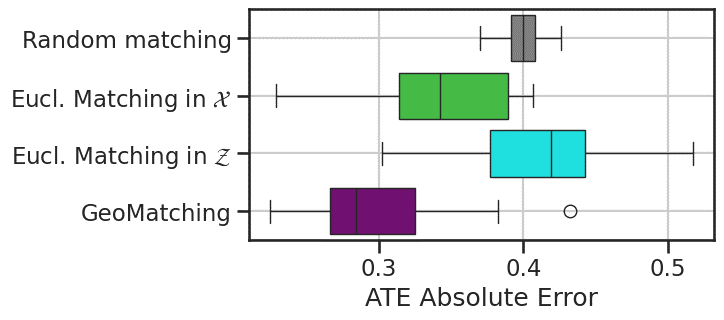}
    \vspace{0.001cm}
    \caption*{(d) ATE Estimation Performance}
    \label{fig:mocal_ate}
  \end{minipage}
  \caption{\textbf{Results for semi-synthetic motion capture data}. (a) Visualization of the latent space and corresponding high-dim confounders after PCA projection. Outliers lie outside the intrinsic data manifold. (b-c) Matched units (geodesic curves) when pairing cases and controls according to Euclidean (baseline) or Riemannian metric (GeoMatching) in the latent space. (d) Average Treatment Effect Absolute Errors (avg across 20 random seeds).}
  \label{fig:mocap}
  \vspace{-0.2cm}
\end{figure} 

Figure~\ref{fig:mocap}(a) shows a visualization of the data in a 2D-latent space using PCA projection, which preserves the geometric/periodic structure of the data. 
In this scenario, we expect GeoMatching’s advantage to be significant for those datapoints in the extremes of the manifold.
%, and performance to be similar for all other datapoints in the middle of the manifold.
%
Indeed, Figures~\ref{fig:mocap}(b) and~\ref{fig:mocap}(c) show that GeoMatching tends to match datapoints in the extremes to other datapoints inside the manifold, whereas Euclidean matching methods fail catastrophically, ignoring the manifold structure and matching several datapoints to the undesired outliers outside the manifold.
All baselines are sensitive to the introduced outliers because they do not take into account data uncertainty; in contrast, GeoMatching is robust against the impact of outliers, which results in better ATE estimation, as shown in Figure~\ref{fig:mocap}(d).\footnote{We note that Figure~\ref{fig:mocap}(d) looks less impressive than Figure~\ref{fig:swissroll}(d) because ATE aggregates performance for all datapoints, and GeoMatching improves performance only for datapoints at the extremes of the manifold.}
These results align with our a priori expectations, as outliers tend to lie in low-density (high-volume) regions which are unlikely to be traversed by geodesics.

\vspace{-0.1cm}
\subsection{Real-world Causality Benchmark Datasets}\label{sec:real-world-exp}
\vspace{-0.2cm}

We evaluate GeoMatching on two real-world datasets. We demonstrate that i) GeoMatching is competitive in more realistic scenarios, and ii) GeoMatching can benefit from semi-supervised settings, where we have access to several unlabelled covariate data, from which only a small subset has treatment and outcome information available.
We consider the Infant Health Development Program Dataset and the Lalonde\footnote{\url{https://users.nber.org/~rdehejia/nswdata2.html}} dataset, see Appendix~\ref{app:datasets} for further details.
%, and ACIC2016 dataset (results for the latter can be found in the Appendix).
%
%We consider the Infant Health Development Program Dataset ($N=747$, $D=25$), Lalonde\footnote{\url{https://users.nber.org/~rdehejia/nswdata2.html}} dataset ($N=747$, $D=25$), and ACIC2016 dataset (results for the latter can be found in the Appendix).

Table~\ref{tab:real-world} summarizes the results on real-world datasets.
%For the IHDP dataset, PCA projection was the one selected for both, Eucl. in Z and GeoMatching, according to ATE performance in model selection step.
For PCA, GeoMatching improves upon all baselines. We also include results for the Isomap projection. In this case, Euclidean matching in $\mathcal{Z}$ is the best approach, slightly above GeoMatching. We hypothesize that this is due to a coarse grid when optimizing the $\sigma$ parameter of the Riemannian metric (only 10 categorical values were explored in a grid search setup, see Appendix for details). In the case of the LaLonde dataset, using a Riemannian metric yield improvements in the Original covariate space, as well as in the PCA and NSM latent representations.

\begin{table}[t!]
    \centering
    \caption{\textbf{ATE absolute error for real-world datasets.}}
    %\caption{Averate Treatment Effect Absolute Errors for real-world datasets}
    \begin{minipage}{\textwidth}
        \centering
        \small
        \caption*{(a) IHDP dataset.}
        % \begin{tabular}{lcc}
        % \toprule
        %  \textbf{ATE abs. errors} & \textbf{PCA} & \textbf{Isomap} \\
        % \midrule
        % Random & $1.269 \pm 0.053$  & $1.269 \pm 0.053$  \\
        % Eucl. in X & $1.126 \pm 0.035$  & $1.126 \pm 0.035$  \\
        % Eucl. in Z & $1.137 \pm 0.027$  & \textbf{1.107 $\pm$ 0.012} \\
        % %NSM & $0 \pm 00$ & $0 \pm 00$ \\
        % \midrule
        % %Geomatching (X) & $1.211 \pm 0.029$ & $1.211 \pm 0.029$ \\
        % GeoMatching (X) & $1.131 \pm 0.036$ & $1.131 \pm 0.036$ \\
        % GeoMatching (Z) & \textbf{1.118 $\pm$ 0.029}  & $1.112 \pm 0.034$ \\
        % %Geomatching (Isomap) & $0 \pm 00$  & $0 \pm 00$ \\
        % %Geomatching (NSM) & $0 \pm 00$ & $0 \pm 00$ \\
        % \bottomrule
        % \end{tabular}
        \begin{tabular}{lcccc}
        \toprule
        & Original ($\mathcal{X}$) & PCA ($\mathcal{Z}$) & Isomap ($\mathcal{Z}$) & NSM ($\mathcal{Z}$) \\ 
        \midrule
        Random & $1.269 \pm 0.053$ & NA & NA & NA \\ 
        Euclidean & \textbf{1.126 $\pm$ 0.035} & $1.137 \pm 0.027$ &  \textbf{1.107 $\pm$ 0.012} & 1.166 $\pm$ 0.093\\ 
        %Mahalanobis & 1.216 $\pm$ 0.063 & 1.146 $\pm$ 0.022 & 1.152 $\pm$ 0.031  \\
        Mahalanobis & 1.216 $\pm$ 0.063 & 1.136 $\pm$ 0.042 & 1.160 $\pm$ 0.051  & 1.207 $\pm$ 0.060\\
        GeoMatching & $1.131 \pm 0.036$ & \textbf{1.118 $\pm$ 0.029} &  $1.112 \pm 0.034$ & \textbf{1.157 $\pm$ 0.094}\\ 
        \bottomrule
        \end{tabular}
    \end{minipage}
    %\hfill
    \vspace{0.3cm}
    \begin{minipage}{\textwidth}
        \centering
        \small
        \caption*{(b) Lalonde dataset.}
        \begin{tabular}{lcccc}
        \toprule
        & Original ($\mathcal{X}$) & PCA ($\mathcal{Z}$) & Isomap ($\mathcal{Z}$) & NSM ($\mathcal{Z}$) \\ 
        \midrule
        Random & 610.1 $\pm$ 300.5 & NA & NA & NA \\ 
        Euclidean & 541.8 $\pm$ 117.6 & 364.9 $\pm$ 219.9 &  1053.2 $\pm$ 921.3 &  811.5 $\pm$ 591.0\\ 
        %Mahalanobis & 858.6 $\pm$ 93.0 & 996.6 $\pm$ 371.0 & \textbf{787.6 $\pm$ 529.54} \\
        Mahalanobis & 809.02 $\pm$ 158.8 & 843.6 $\pm$ 438.1 & \textbf{766.5 $\pm$ 468.4} & 948.8 $\pm$ 448.7 \\ 
        GeoMatching & \textbf{485.6 $\pm$ 319.3}  & \textbf{363.2 $\pm$ 122.6} &  809.6 $\pm$ 456.5 & \textbf{561.4 $\pm$ 396.3}\\ 
        \bottomrule
        \end{tabular}
        % \begin{tabular}{lcc}
        % \toprule
        %  \textbf{ATE abs. errors} & \textbf{PCA} & \textbf{Isomap} \\
        % \midrule
        % Random & 610.1 $\pm$ 300.5 & 610.1$\pm$ 300.5   \\
        % Eucl. in X & 541.8 $\pm$ 117.6   & 541.8 $\pm$ 117.6  \\
        % Eucl. in Z &  364.9 $\pm$ 219.9 &  1053.2 $\pm$ 921.3 \\
        % %NSM & $0 \pm 00$ & $0 \pm 00$ \\
        % \midrule
        % Geomatching (X) & 485.6 $\pm$ 319.3  & \textbf{485.6 $\pm$ 319.3} \\
        % Geomatching (Z) & \textbf{363.2 $\pm$ 122.6} & 809.6 $\pm$ 456.5 \\
        % %Geomatching (Isomap) & $0 \pm 00$  & $0 \pm 00$ \\
        % %Geomatching (NSM) & $0 \pm 00$ & $0 \pm 00$ \\
        % \bottomrule
        % \end{tabular}
    \end{minipage}
    %\vspace{-0.4cm}
    \label{tab:real-world}
\end{table}

\vspace{-0.1cm}
\section{Conclusion}\label{sec:conclusion}
\vspace{-0.2cm}

In this paper, we adopt a principled differential geometric approach for causal inference.
%
% We propose to empower the latent representation/data manifold with a Riemannian metric, considering Riemannian distances along the manifold for matching.
%
We introduce GeoMatching, a novel matching method for treatment effect estimation that accounts for geometry structure and uncertainty of the confounders. The idea is to match cases and controls based on a latent Riemannian metric instead of a Euclidean metric, such that distances are more meaningful along the data manifold.
We learn a non-parametric metric space by constructing a smoothly changing kernel-based deterministic  metric that induces a Riemannian manifold. We empirically show that GeoMatching outbeats other baselines for treatment effect estimation, and remains compatitive for increasing input dimensionality, in the presence of outliers, and semi-supervised scenarios.

% We measure distances along the covariates data manifold in a differential-geometric principled manner, leveraging the geometric information from the data manifold.

\paragraph{Limitations}
We note that the manifold learning stage, i.e., projecting the data in a way that preserves geometric structure, as well as the computational cost from computing geodesics can be improved.
In general, neither PCA nor Isomap are probabilistic approaches, and neither of them are guaranteed to preserve geometric/topological structure. 
If the GeoMatching is performed in a latent representation that destroys geometric structure, treatment effects could be arbitrarily biased, potentially leading to negative societal impact. This is true for any matching method on latent representations.
Computationally, the bottleneck is how to evaluate geodesics, which can be speed-up by minimizing a discretized length of the curve using graph information~\cite{chen_metrics_2018}, or using approximate Riemannian metrics~\cite{arvanitidis_prior-based_2022}.
We note that GeoMatching is a framework rather than a closed-form procedure, allowing for enhancements in learning better latent representations or accelerating the computation of geodesics to be seamlessly integrated.

% \cite{chen_metrics_2018}: How to compute geodesic distances by learning a parametrized model for the curve, and minimizing discretized length of the curve.

\paragraph{Broader Impact} Geometry is an important inductive bias in ML and has been shown to yield benefits in diverse ML applications such as classification, interpolation, model training, and synthetic data generation. To the extend of our knowledge, this work is the first to look into the benefit of geometry-aware methods for causal inference.
Estimation of treatment effects using differential geometry is a promising avenue for future research, given that causality often entails geometric manifold structures naturally~\citep{dominguez-olmedo_data_2023}.
As future work, an interesting research direction would be to combine GeoMatching with causal discovery methods to inform the construction of the latent space. Instead of learning the entire manifold at once, one could learn all independent mechanisms separately and explicitly parameterize the manifold for computing geodesics, improving computational efficiency and accuracy.
Another interesting research direction would be to account for geometry broadly in other causal inference methods, including meta-learners, synthetic controls or propensity-score weighting methods, leveraging Riemannian geometry instead of Euclidean spaces.
%As future work, we plan to explore geometry-aware synthetic controls, as well as geometry-aware propensity-score weighting methods, leveraging Riemannian geometry instead of Euclidean spaces.
%
%
% TODO: Add 
%Enables incorporation of suitable inductive biases in the TE estimation methods.
%

%We note that GeoMatching is a framework rather than a closed-form procedure, so any improvements on learning better latent representations or speeding up computation of geodesics can be armored within the framework.%, as long as the latent representation preserves geometric structure of the confounders.

% Acknowledgments---Will not appear in anonymized version
\section*{Acknowledgments}
We thank Javier Zazo for continuous discussions and technical support. We also thank Aditya Nori for early discussions to the initial idea of this research.
% We gratefully acknowledge Javier Zazo for his insightful discussions on the technical details, which significantly enhanced the depth and rigor of this study. We also thank Aditya Nori for early discussions and his invaluable support to the initial idea of this research. 

\bibliography{geomatching_paper}

\appendix

\newpage
\section{Concepts of Riemannian Geometry}\label{app:riemannian}

In this Section, we review key concepts of Riemannian geometry that will be useful for our work.

%\textbf{Manifold}.
\begin{definition}
(Manifold). A $d$-dimensional smooth manifold $\mathcal{M}$ is a topological space which locally resembles the Euclidean space $\mathbb{R}^d$ and has a smooth structure.
\end{definition}

%\textbf{Riemannian manifold}.
\begin{definition}
(Riemannian Manifold). A Riemannian manifold is a differentiable (smooth) manifold $\mathcal{M}$ provided with a Riemannian metric tensor $\G$.
\end{definition}
%A smooth manifold equipped with a \textit{Riemannian metric}.
% $\mathbb{G}:\;\mathcal{M} \rightarrow \mathcal{S}^k_{++}$ is a smooth function that assigns a symmetric positive definite matrix to any point in $\mathcal{M}$.

\begin{definition}
(Riemannian metric). A Riemannian metric $\G$ on a manifold $\mathcal{M}$ is a smooth function that assigns a symmetric positive definite matrix to any point $z \in \mathcal{M}$.
\end{definition}
A Riemannian metric defines at each point $z$ a smoothly varying inner product in the tangent space $T_z\mathcal{M}$. The inner product is defined as:
\begin{equation}
      \langle z_1,z_2 \rangle_z = z_1^{\mathrm{T}}\G(z)z_2,
\end{equation}
where $z_1, z_2 \in T_z\mathcal{M}$ and $z \in \mathcal{M}$.
Intuitively, a Riemannian metric defines an infinitesimal notion of distance on the manifold $\mathcal{M}$. The length of a smooth curve $\gamma: [0,1] \rightarrow \mathcal{M}$ is then defined as:
\begin{equation}
    \mathrm{Length}(\gamma) = \int_0^1 \sqrt{\dot{\gamma}(t)^T \G(\gamma(t)) \dot{\gamma}(t) dt},
\end{equation}
where $\dot{\gamma}(t) = \frac{d}{dt}\gamma(t)$ denotes the velocity of the curve. The distance between two points on the manifold $\mathcal{M}$ is defined as the length of the shortest curve connecting them.

\begin{definition}
(Geodesic). A geodesic curve between two points $x_1$ and $x_2$ is a length-minimising curve connecting the two points:
\begin{equation}
    \gamma_{\mathbf{G}} = \underset{\gamma}{\mathrm{argmin}} \;\mathrm{Length(\gamma)}, \qquad \gamma(0)=x_1, \gamma(1) = x_2.
\end{equation}
\end{definition}
We call \textit{Riemannian} or \textit{geodesic distance} the length of the geodesic curve, according to the underlying Riemannian metric $\mathbf{G}$.

\begin{definition}
(Riemannian distance). The Riemannian distance $d_{\G}(r,s)$ between two points $r,s \in \mathcal{M}$ on a Riemannian manifold $(\mathcal{M, \mathbb{G}})$ is defined as the infimum of the length of all smooth curves  $\gamma: [0,1] \rightarrow \mathcal{M}$ connecting $r$ and $s$,
\begin{equation}
    d_{\G}(r,s) = \inf \{ \mathrm{Length(\gamma) \given \gamma(0)=r,\, \gamma(1)=s}. \}
\end{equation}
\end{definition}

\begin{definition}
    (Magnification factor) The magnification factor -- also called Riemannian volume -- measures the magnitude of the local distortion of space at location $p \in \mathcal{M}$, i.e., $V_{\G}(p) := \sqrt{\mathrm{det}\, \G(p)}$.
\end{definition}
Curves that traverse regions with high magnification factor will tend to have larger lengths (it can also be seen as how much energy is needed to traverse such space). Geodesics or shortest paths tend to avoid regions with high values of magnification factor (which reflect high uncertainty away from the observed data manifold).

% \begin{definition}
% (Pullback metric) Let $f: \mathcal{Z} \rightarrow \mathcal{X}$ be a smooth mapping between two smooth manifolds $\mathcal{Z},\mathcal{X}$, and let $\mathcal{X}$ be equipped with a Riemannian metric $\G'$. The metric $\G'$ can be "pulled back" to $\mathcal{Z}$ via the \textit{pullback metric}:
% \begin{equation}
%     \G(z) := J_f(z)^T \G'(f(z)) J_f(z)
% \end{equation}
% where $J_f(z)$ is the Jacobian of $f$ at location $z \in \mathcal{Z}$. The pullback metric enables transferring the infinitesimal notion of distance of manifold $\mathcal{X}$ into manifold $\mathcal{Z}$.
% \end{definition}

% \paragraph{Conformal metrics, conformal equivalence} When metrics are identical ``up to scale'', i.e.,
% \begin{equation}
%     \exists \lambda: \mathcal{M} \rightarrow (0,\infty) s.t. G'(p) = \lambda(p) G(p) \forall p \in \mathcal{M}.
% \end{equation}
% We then say that $G'$ and $G$ are  conformally equivalent, and $\lambda$ is referred to as the \textit{conformal factor}.

\newpage

\section{Further Information about Datasets}\label{app:datasets}

\paragraph{Synthetic Data Generation for the Swissroll Dataset.}

In Section~\ref{subsec:swissroll}, we generate a three-dimensional swissroll in a space of higher dimensionality $D \in \{3, 5, 10, 15, 25, 50, 100\}$. The data generation process for a specific datapoint $(X,T,Y)$ is as follows:
\begin{eqnarray}
R & \sim & \mathcal{U}([1.5 \pi, 4.5 \pi]) \\
X_1 & = & R \cos(R) \\
X_2 & = & R \sin(R) \\
X_3 & \sim & \mathcal{U}([0, 8]) \\
X_4, \ldots, X_D & \sim & \mathcal{N} \left(0, \sigma_x \mathbf{I_{D-3}} \right) \\
T & \sim & \mathrm{Bernoulli} \left( \mathrm{sigmoid}(3 \pi) R \right) \\
\mu_0(x) & = & -\frac{1}{2} x + \frac{3}{2} \pi \\
\mu_1(x) & = & -\frac{3}{2} x + \frac{9}{2} \pi\\
Y(0) & \sim & \mathcal{N}( \mu_0(R - \frac{3}{2} \pi), \sigma_y \mathbf{I}) \\
Y(1) & \sim & \mathcal{N}( \mu_1(R - \frac{3}{2} \pi), \sigma_y \mathbf{I}) \\
Y & = & (1-T) Y(0) + T Y(1), 
\end{eqnarray}
where $R$ is the intrinsic manifold where the data lives, $\sigma_x$ is the standard deviation for the additional uninformative input dimensions, and $\sigma_y$ is the noise standard deviation for outcome $Y$. Note that the results we report in the main text assume a 2D latent representation, we are essentially matching in the $(X_1,X_2)$ space, filtering out the rest of input covariates. While for this specific toy, we could have mapped the original space $\mathcal{X}$ into a 1D Euclidean space, in which case the Riemannian distance would be equivalent to the Euclidean distance, this is not always feasible. Thus, we deliberately project the original confounders into a 2D latent space to illustrate the more generic situation in which the data manifold cannot be mapped to a Euclidean space.
%
%we embed the three-dimensional swissroll in a space of higher dimensionality $D \in \{3, 5, 10, 15, 25, 50, 100\}$, by adding additional noisy (irrelevant) input dimensions.

\paragraph{Semi-synthetic Data Generation for the CMU Mocap Dataset.}

We consider real-world covariates from the CMU Motion Capture Database\footnote{\url{http://mocap.cs.cmu.edu}} (motion 16 from subject 22), and we synthetically generate the treatment and outcome variables.
We chose human motion data because it has been used in previous works to illustrate the impact of
geometry on downstream applications, as latent spaces are expected to be cylindrical or toroidal/doughnut-shaped~\citep{urtasun_topologically-constrained_2008, tosi_metrics_2014}.
%Each observation corresponds to a human pose as acquired by a marker-based motion capture system.
%
%\textit{Injection of outliers}: To make the data more challenging, we intentionally perturb some time frames to simulate outlier confounders in the dataset (i.e., players jumping in a weird, unnatural manner, or motion capture sensors being defective).

Regarding the covariates, for each random seed we inject two outliers by randomly selecting a motion capture frame of each side of the data manifold (corresponding to frames with feet close to the ground, one treated and one control unit). We then obtain a perturbed frame by linear interpolation of these two covariates, in order to get a realistic motion capture outside of the main data manifold. The data generation process for the generation of $(T,Y)$ is as follows:
\begin{eqnarray}
    R & \sim & \mathcal{U}([0, 6 \pi]) \\
    T & \sim & \mathrm{Bernoulli} \left( \mathrm{sigmoid}(z_1) \right) \\
    \mu_0(x) & = & 0.4 \sin(x + \frac{1}{2}) + \frac{1}{2} \\
    \mu_1(x) & = & 0.4 \sin(x + \frac{1}{2}) + \frac{1}{2}\\
    Y(0) & \sim & \mathcal{N}( \mu_0(R), \sigma_y \mathbf{I}) \\
    Y(1) & \sim & \mathcal{N}( \mu_1(R), \sigma_y \mathbf{I}) \\
    Y & = & (1-T) Y(0) + T Y(1), 
\end{eqnarray}

\paragraph{IHDP Dataset.} The Infant Health and Development Program (IHDP) is a randomized controlled study conducted between 1985 and 1988, designed to evaluate the effect of home visit from specialist doctors on the cognitive test scores of premature infants.~\citep{hill_bayesian_2011}. We chose this dataset because it is commonly used as a benchmark dataset in the causality community, as it allow us to assess performance against ground truth
values~\cite{hill_bayesian_2011, curth_doing_2021}.
We use the synthetic version of IHDP introduced by~\cite{louizos_causal_2017}, as it allow us to assess performance against ground truth values. We preprocess the data according to the recommendations in~\cite{curth_doing_2021}.

\paragraph{LaLonde Dataset.} The LaLonde dataset is a dataset from econometrics which looks at the impact of training program policies in employee's salary~\citep{lalonde_evaluating_1986, dehejia_causal_1998}. The treatment is whether the participant attended a job training program, the outcome is the earning in 1978, and pre-treatment covariates include 6 covariates capturing demographic information. The dataset contains interventional (RCT) and observational data. Replicating the experimental protocol from~\cite{kuang_estimating_2017}, we consider treated units from the RCT, and controls from observational data. Thanks to the complete RCT data, we can estimate a ground truth value for the ATE. Our goal is then to estimate the ATE using controls from observational data, which requires handling confounding bias/correcting for distribution imbalance between cases and controls.

\newpage
\section{Further Empirical Results}\label{app:empirical}

\subsection{Additional Results for Datasets considered in Main Manuscript}

\begin{figure}[h!]
    \centering
    \includegraphics[width=\linewidth]{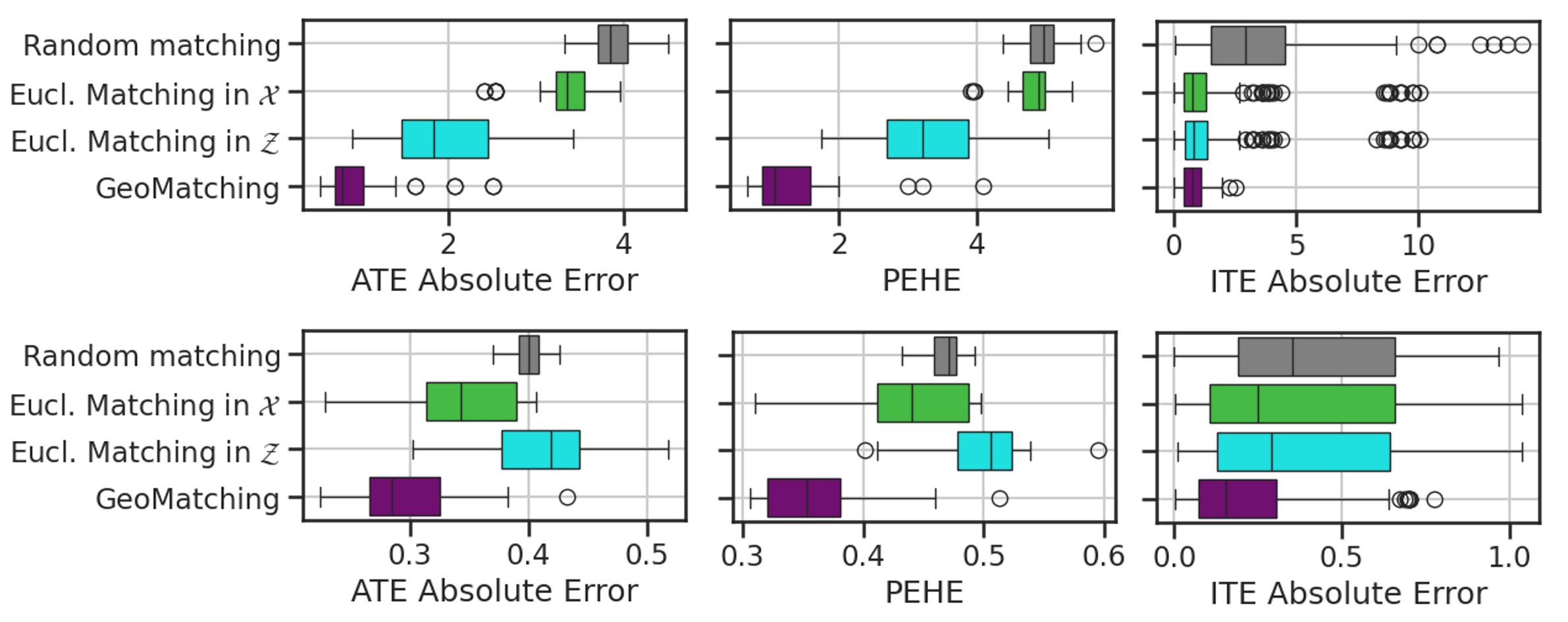}
    \caption{\textbf{Additional Results for (Semi)-Synthetic Datasets}. (1st row) Swissroll data; (2nd row) Motion Capture data. 1st column is the same as in the main text to facilitate comparison; 2nd column corresponds to Precision in Estimation of Heterogeneous Treatment Effects (PEHE) averaged over all seeds, and 3rd column shows Individual Treatment Effect (ITE) Absolute Error for a single seed.}
    \label{fig:app-pehe-toys}
\end{figure}

\begin{table}[h!]
    \centering
    %\caption{\textbf{PEHE for real-world datasets.}}
    \caption{\textbf{Precision in Estimation of Heterogeneous Effect (PEHE).}}
    \begin{minipage}{\textwidth}
        \centering
        \small
        \caption*{(a) IHDP dataset.}
        \begin{tabular}{lcccc}
        \toprule
        & Original ($\mathcal{X}$) & PCA ($\mathcal{Z}$) & Isomap ($\mathcal{Z}$) & NSM ($\mathcal{Z}$) \\ 
        \midrule
        Random & 1.613 $\pm$ 0.083 & NA & NA & NA \\ 
        Euclidean & 1.452 $\pm$ 0.016 & \textbf{1.399 $\pm$ 0.024} &  1.422 $\pm$ 0.034 & 1.448 $\pm$ 0.113 \\ 
        Mahalanobis & 1.543 $\pm$ 0.048 & 1.433 $\pm$ 0.035 & 1.409 $\pm$ 0.043 & 1.506 $\pm$ 0.087 \\
        GeoMatching & \textbf{1.433 $\pm$ 0.023} & 1.416 $\pm$ 0.056 &  \textbf{1.390 $\pm$ 0.042} & \textbf{1.438 $\pm$ 0.107} \\ 
        \bottomrule
        \end{tabular}
        % \begin{tabular}{lcccc}
        % \toprule
        % & Original ($\mathcal{X}$) & PCA ($\mathcal{Z}$) & Isomap ($\mathcal{Z}$) & NSM ($\mathcal{Z}$) \\ 
        % \midrule
        % Random & $1.346 \pm 0.045$ & NA & NA & NA \\ 
        % Euclidean & 1.227 $\pm$ 0.028 & 1.105 $\pm$ 0.029 &  1.231 $\pm$ 0.057 & $1.448 \pm 0.113$ \\ 
        % Mahalanobis & \textbf{1.091 $\pm$ 0.045} & 1.159 $\pm$ 0.047 & \textbf{1.132 $\pm$ 0.051} & $1.506 \pm 0.087$ \\
        % GeoMatching & $1.202 \pm 0.035$ & \textbf{1.094 $\pm$ 0.022} &  $1.162 \pm 0.054$ & \textbf{1.438 $\pm$ 0.107}\\ 
        % \bottomrule
        % \end{tabular}
    \end{minipage}
    %\hfill
    %\vspace{0.1cm}
    \begin{minipage}{\textwidth}
        \centering
        \small
        \caption{For Lalonde dataset, no true ITE available, so we cannot compute PEHE.}
        % \begin{tabular}{lcccc}
        % \toprule
        % & Original ($\mathcal{X}$) & PCA ($\mathcal{Z}$) & Isomap ($\mathcal{Z}$) & NSM ($\mathcal{Z}$) \\ 
        % \midrule
        % Random & 610.1 $\pm$ 300.5 & NA & NA & NA \\ 
        % Euclidean & 541.8 $\pm$ 117.6 & 364.9 $\pm$ 219.9 &  1053.2 $\pm$ 921.3 & $000.0 \pm 000.0$\\ 
        % Mahalanobis & 809.02 $\pm$ 158.8 & 843.6 $\pm$ 438.1 & \textbf{766.5 $\pm$ 468.4} & $000.0 \pm 000.0$ \\ 
        % GeoMatching & \textbf{485.6 $\pm$ 319.3}  & \textbf{363.2 $\pm$ 122.6} &  809.6 $\pm$ 456.5 & $000.0 \pm 000.0$\\ 
        % \bottomrule
        % \end{tabular}
    \end{minipage}
    \label{tab:real-world-PEHE}
\end{table}

\subsection{Results for U-shaped Strip Synthetic Toy}

We generate an additional synthetic toy consisting of a U-shaped Strip as follows:  
\begin{figure}[ht]%[htbp]
  \centering
  \begin{minipage}{.3\textwidth}
  \centering
  \vspace{-3.5cm}
  \begin{tikzpicture}[->,>=stealth',shorten >=1pt,auto,node distance=2cm,
                      thick,main node/.style={circle,draw,font=\sffamily\footnotesize\bfseries}] %#fill=blue!20

    \node[main node] (1) {$X$};
    \node[main node, fill=gray!20] (2) [below of=1] {$Z$};
    \node[main node] (3) [below left of=2] {$T$};
    \node[main node] (4) [below right of=2] {$Y$};

    \path[every node/.style={font=\sffamily\small}]
      (2) edge node [left] {} (1)
          edge node [left] {} (3)
          edge node [right] {} (4)
      (3) edge node [right] {} (4);
  \end{tikzpicture}
  \caption*{(a) True Causal Graph}
  \label{fig:u-shape-toy-a}
\end{minipage}
\begin{minipage}[b]{0.3\textwidth}
    \includegraphics[width=\textwidth]{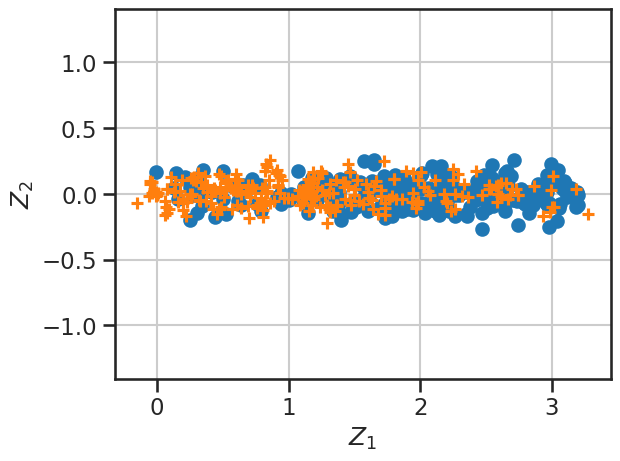}
    \caption*{(b) Latent Representation $Z$}
    \label{fig:u-shape-toy-b}
  \end{minipage}
  \begin{minipage}[b]{0.3\textwidth}
    \includegraphics[width=\textwidth]{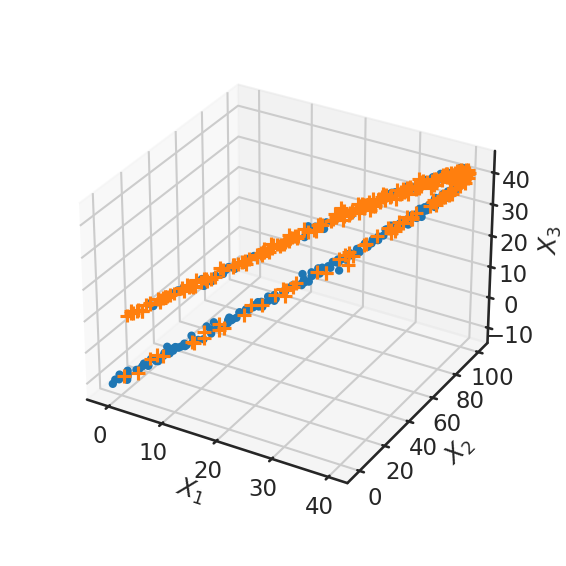}
    \caption*{(c) Observational Data $X$}
    \label{fig:u-shape-toy-c}
  \end{minipage}
  \\
  \begin{minipage}[b]{0.3\textwidth}
    \includegraphics[width=\textwidth]{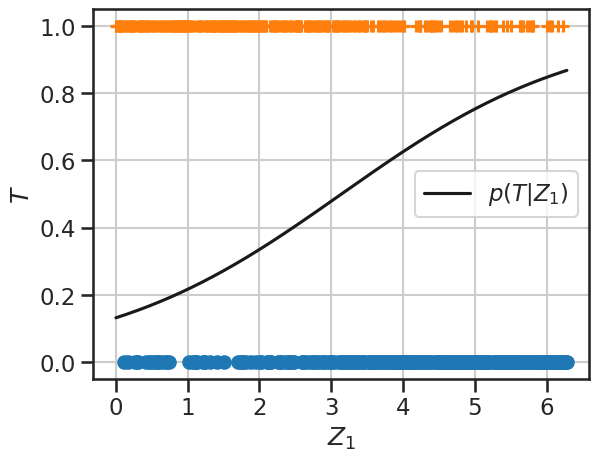}
    \caption*{(d) Treatment Assignment $T$}
    \label{fig:u-shape-toy-d}
  \end{minipage}
  \begin{minipage}[b]{0.3\textwidth}
    \includegraphics[width=\textwidth]{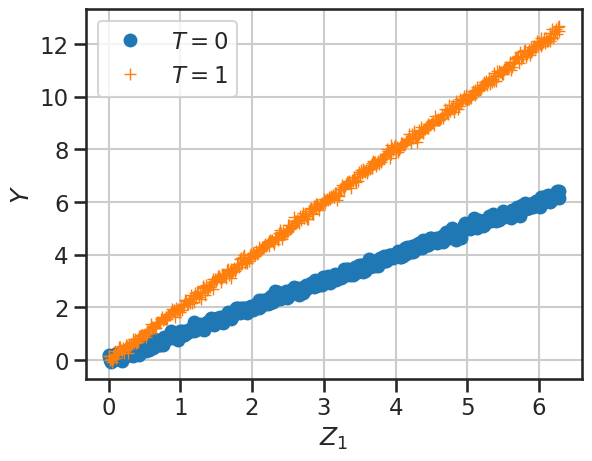}
    \caption*{(e) Treatment Response $Y$}
    \label{fig:u-shape-toy-e}
  \end{minipage}
  \begin{minipage}[b]{0.3\textwidth}
    \includegraphics[width=\textwidth]{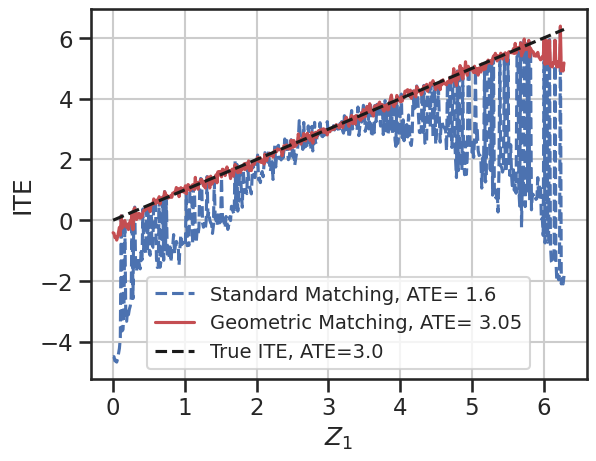}
    \caption*{(f) ATE Estimation}
    \label{fig:u-shape-toy-f}
  \end{minipage}
  % \begin{minipage}[b]{0.3\textwidth}
  %   \includegraphics[width=\textwidth]{figures/toy1/toy17.png}
  %   \caption{Geometric Matching}
  %   \label{fig:subfig6}
  % \end{minipage}
  \caption{\textbf{Results for U-shaped Strip Synthetic Toy.} This proof-of-concept toy shows that accounting for data geometry improves estimation of treatment effects. %We use 20 neighbors for both matching methods.
  }
  %\label{fig:u-shape-toy}
\end{figure}

\begin{eqnarray}
Z & \sim & \mathcal{U}([0,2\pi]) \\
X \given Z & \sim & \mathcal{N}(f(Z),\sigma^2_x I) \\
T \given Z & \sim & \mathcal{N}(\mathrm{sigmoid}(Z_1;k,z_0),\sigma^2_x I) \\
Y \given T,Z & \sim & T\mathcal{N}(2 Z_1; \sigma^2_Y) + (1-T)\mathcal{N}(Z_1; \sigma^2_Y) \\
ITE \given Z & \sim & \mathcal{N}(Z_1; 2\sigma^2_Y)
\end{eqnarray}

\paragraph{Increasing dimensionality of observational space for U-shape synthetic toy}

Similar to Figure~\ref{fig:swissroll}(e) in the main text, we embed the U-shape strip in a high-dimensional space by adding additional uninformative input dimensions. Standard matching degrades as we increase the input dimensionality, whereas GeoMatching performance remains stable.

% \begin{eqnarray}
% Z_1 & \sim & \mathcal{U}([0,2\pi]) \\
% Z_2, \ldots, Z_D & = & 0 \\
% X \given Z & \sim & \mathcal{N}(Z,\sigma^2_x I) \\
% T \given Z & \sim & \mathcal{N}(\mathrm{sigmoid}(Z_1;k,z_0),\sigma^2_x I) \\
% Y \given T,Z & \sim & T\mathcal{N}(2 Z_1; \sigma^2_Y) + (1-T)\mathcal{N}(Z_1; \sigma^2_Y) \\
% ITE \given Z & \sim & \mathcal{N}(Z_1; 2\sigma^2_Y)
% \end{eqnarray}

\begin{figure}[ht]
    \centering
    \includegraphics[width=0.45\textwidth]{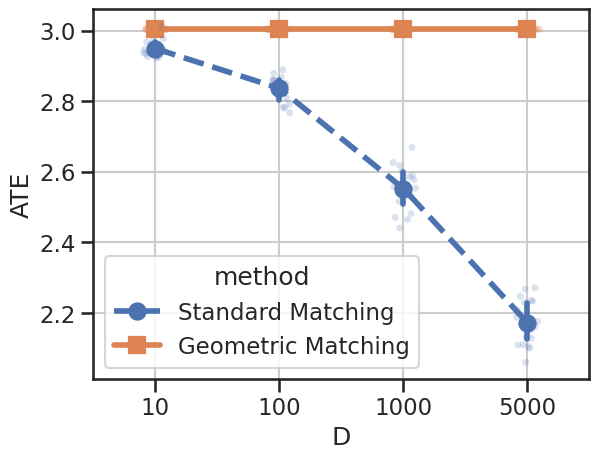}
    \caption{\textbf{Impact of dimensionality $D$ on TE estimation.} As $D$ increases, matching all covariates perfectly becomes impossible. %This example does not modify the geometry, first dimension $Z_1$ corresponds to the 
    }
    \label{fig:u-shape-toy-highdim}
\end{figure}

\subsection{Sensitivity Analysis w.r.t Hyperparameters}

\begin{figure}[ht]
    \centering
    \includegraphics[width=\linewidth]{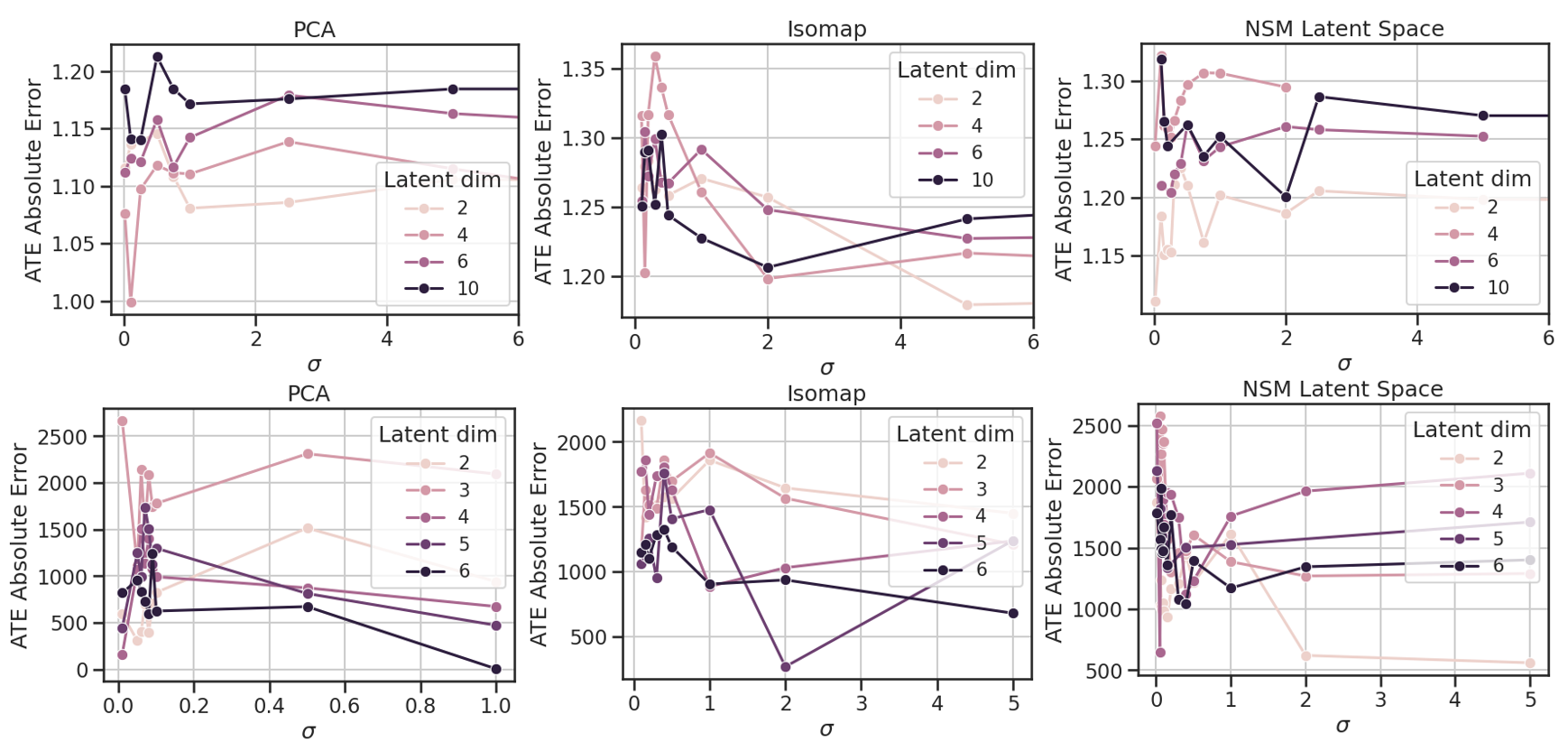}
    \caption{\textbf{Sensitivity Analysis of GeoMatching w.r.t different Hyperparameters}. We consider variability for a single random seed in ATE Absolute Error performance w.r.t different latent representations (PCA, Isomap, and NSM latent space), latent dimensionality, and $\sigma$, the parameter that controls the curvature of the Riemannian manifold.}
    \label{fig:sensitivity}
\end{figure}

Figure~\ref{fig:sensitivity} depicts variation of TE estimation performance as we vary the two hyperparameters considered in this paper: the dimensionality of the latent representation and the curvature $\sigma$ of the Riemannian manifold.
GeoMatching is sensitive to the selected hyperparameters (manifold learning is a non-trivial task in general). We selected hyperparameters based on ATE performance in a separate validation set. Worth noting is that, even if the grid search we ran was quite coarse (due to limited compute), we were still able to observe improvements empirically. We hypothesize that results would be much better with a finer grid, or using Bayesian optimization to dynamically optimize parameter $\sigma$, which controls the curvature of the Riemannian manifold.

\newpage

\section{Bias Decomposition}\label{app:bias_decomposition}

In this section, we conduct a theoretical analysis based on the decomposition of the estimation bias for GeoMatching.
In the following, we assume that the \textit{manifold hypothesis} and \textit{geometric faithfulness} assumptions stated in Section~\ref{sec:motivation} hold, resulting in a \textit{well-specified} representation of covariates.
The geometric faithfulness assumption ensures that points close to each other in the latent manifold will have similar treatment effects.
%
%We then prove that GeoMatching is optimal (i.e., minimizes extrapolation bias) if the mapping function from the latent space to the outcome is a distance-preserving transformation (isometry).
By decomposing the estimation bias, we show that the Riemannian metric is optimal (minimizes extrapolation bias) when the mapping function from covariate to outcome space is an isometry.

Let us consider covariates $X_\star \in \mathbb{R}^d$ of a treated individual ($T_\star = 1$). 
Let $\mu_t(x)=\mathbb{E}[Y(t) \given X=x]$ denote the expected outcome under treatment $T=t$.
Our goal is to estimate the individual treatment effect $\mathrm{ITE}(X_\star) = |\mu_1(X_\star) - \mu_0(X_\star)|$, where $\mu_0(X_\star)$ is unobserved.
%
%As in previous sections, let $X_\star \in \mathbb{R}^d$ be the covariates of an individual exposed to a treatment $T_\star = 1$. Let $\mu_1(x)=\mathbb{E}[Y(1) \given X=x]$ and $\mu_0(x)=\mathbb{E}[Y(0) \given X=x]$ denote the expected outcome with or without treatment. Our object of interest is the individual treatment effect $\mathrm{ITE}(X_\star) = |\mu_1(X_\star) - \mu_0(X_\star)|$.
%
%As stated in Equation~\ref{eq:problem_formulation}, matching requires computing the distance $d(X_\star, X_j)$ between treated unit $X_\star$ and control unit $X_j$.
%
Given a fixed distance metric $d(\cdot,\cdot)$ in covariate space, we can solve 
$j^{\dagger} = \underset{j \in [m]}{\text{argmin}} \; d(X_{\star}, X_j)$ among a pool of $m$ controls, 
%Equation~\eqref{eq:problem_formulation}
and build an ITE matching estimator:
%
%Intuitively, we want to choose the distance that minimizes the \textit{extrapolation bias} of the matching estimator:
\begin{eqnarray}
    \widehat{\mathrm{ITE}}(X_\star; d) = |\mu_1(X_\star) - \mu_0(X_{j^\dagger(d)})| = | \underbrace{\mu_1(X_\star) - \mu_0(X_\star)}_{\mathrm{ITE}(X_\star)} + \underbrace{\mu_0(X_\star) - \mu_0(X_{j^\dagger(d)})}_{\mathrm{Bias}(X_\star; d)} |,
\end{eqnarray}
where $\mathrm{Bias}(X_\star; d)$ is the \textit{extrapolation bias} or cost to pay from using covariate $X_{j^\dagger(d)}$ instead of $X_\star$.

% In an ideal scenario, we can estimate the treatment effect perfectly, i.e., $\mathrm{Bias}(X_\star; d) = 0$, by finding a control unit $X_{j^\dagger(d)}$ s.t. $d(X_\star, X_{j^\dagger(d)}) = 0$.
% %
% \textit{Perfect matching} methods aim to recover pre-treatment covariates perfectly, but 
% %
% % This corresponds to \textit{perfect matching}, whose goal is to recover pre-treatment covariates perfectly:
% % \begin{equation}
% %     \sum^m_{j=1} w_j X_j = X_\star. %X_{j^{\dagger}} = X_\star.
% % \end{equation}
% %
% in practice, it is often unfeasible to find perfect matches.
% %
% This becomes a bigger challenge as dimensionality and noise of the covariates increases.

When matched units are imperfect,
%, i.e., $d(X_{\star}, X_{j^\dagger}) \geq 0$,
matching estimators need to extrapolate, incurring in a non-zero extrapolation bias:
%$|\mu_0(X_{\star}) - \mu_0(X_j) | \geq 0$. 
\begin{equation}
    d(X_{\star}, X_{j^\dagger(d)}) \geq 0 \quad \Rightarrow \quad \mathrm{Bias}(X_\star; d) = |\mu_0(X_{\star}) - \mu_0(X_{j^\dagger(d)}) | \geq 0.
\end{equation}
For example, if the expected outcome without treatment is linear $\mu_0(x) = \beta x$, we incur a bias $d(X_{\star}, X_{j^\dagger(d)}) \beta$ directly proportional to the distance to the match in covariate space.
We say a distance is optimal if it minimizes the bias of the matching estimator: \begin{equation} d^\dagger = \underset{d}{\text{argmin }} \mathrm{Bias}(X_\star; d). \end{equation}
GeoMatching pairs treated and control units along the Riemannian manifold by minimizing: \begin{equation} j^{\star}
= \underset{j \in [m]}{\text{argmin }} d_{\mathcal{R}}(X_{\star}, X_j). \end{equation} Ideally, we would like to match
treated and control units such that the extrapolation bias gets minimized: \begin{equation} j^{\star} = \underset{j \in
[m]}{\text{argmin }} | \mu_0(X_{\star}) - \mu_0(X_j) |. \end{equation}
Thus, GeoMatching is optimal, i.e., minimizes the extrapolation bias, when the expected outcome function $\mu_0$ is an
isometry, i.e., a distance preserving transformation from latent to outcome space.

We note that the provided analysis elucidates which condition needs to be fulfilled for the extrapolation bias to be minimized, and holds for any distance metric $d$, including Riemannian distance (but it is not specific to the Riemannian distance).
What is specific to GeoMatching are the two key underlying assumptions stated in Section~\ref{sec:motivation}.

The analysis in this section studies the error due to \textit{the choice of distance metric} given a fixed representation of covariates, i.e., which distance metric minimizes extrapolation bias assuming a well-specified representation.
In contrast to our case, \cite{clivio_towards_2024} study the error due to the \textit{choice of the representation} given a fixed distance metric. A future research direction would be to provide a full theoretical characterization of GeoMatching by leveraging the theoretical results of \cite{clivio_towards_2024}.

\newpage
\section{Computational Impact of GeoMatching}\label{sec:computational}

% The cost of computing a single geodesic scales linearly with the latent dimensionality

Solving an ODE, to compute a single geodesic distance, scales linearly with the intrinsic dimensionality of the latent space~\citep{kramer_probabilistic_2022}. We need to solve one ODE to compute each pairwise distance. The current implementation of GeoMatching (which we will make publicly available), relies on naïve nearest neighbour search, which scales linearly with the latent dimensionality, and quadratically with the number of datapoints.

%\textbf{Further implementation details}:
We leverage the Stochman public library for computation of geodesics. We compute geodesics using CPUs and sequentially, which is unsurprisingly not very fast (each experiment would take around one night to run, we launched a Wandb grid sweep in a cluster of 1000 nodes in parallel).

We note that our work does not focus on computational efficiency, but on a conceptual and methodological improvement of matching methods. Making GeoMatching computationally efficient is a promising research direction, which can be done in several regards: i) computing geodesics faster by constraining the space of Riemannian metrics to learn~\citep{arvanitidis_prior-based_2022}, ii) leveraging extensions to nearest neighbor search such as Locality Sensitive Hashing to scale sub-linearly instead of quadratically, iii) exploiting hardware improvements, using GPUs and parallel computation.% for geodesics.

\newpage
\section{Frequently Asked Questions}

\subsection*{Can we apply GeoMatching to out of sample data?}

Computing TEs for an unknown observation (very different from any sample seen so far in the train set) is problematic
for any matching method, as it violates the overlap assumption~\citep{rubin_causal_2005}. All reported results correspond to in-distribution performance; out-of-distribution is out of scope.

\subsection*{Does causal structure always imply geometric structure?}

Some geometrical structures could be implied by some causal structure but not all causal structure would reflect on the structure of the data. We note that geometric structure alone might not be enough to recover causal structure; this relates to
identifiability of the causal graph. The key assumption of this work is that  causal mechanisms most often constraint observations to be sparse/non i.i.d. (e.g., given a set of patient covariates, we would never get a realistic patient record if sampling from each covariate marginally) and smooth (e.g., there exists a smooth variation in patients vitals/measurements).

\subsection*{What is the role of the manifold hypothesis in Causal Inference?}

The manifold hypothesis states that the high-dimensional covariates $X$ lie near a low-dimensional (non-linear) manifold embedded in the original input space. Such hypothesis is not commonly discussed nor exploited in the causality field, despite being widely accepted within the machine learning community. This work aims to bring awareness to and exploit the manifold hypothesis in the context of causal inference, bridging the gap between both communities.

%the manifold hypothesis is less commonly discussed/exploited in causality despite being widely accepted in the ML community. Our work brings such new perspective and has the potential to influence the CI sub-community

There is a plethora of arguments supporting the existence of low-dim structure in high-dim data~\citep{loaiza-ganem_deep_2024}, which could directly benefit causal inference. First, the manifold hypothesis implies two properties that are commonly observed in real-world data: data sparsity in input space, and a notion of smoothness in
observations. Second, theoretical works have shown that learning complexity scales exponentially with intrinsic dimensionality, and yet, modern algorithms are still successful at learning low-dim representations~\citep{bengio_representation_2013}, providing strong implicit justification for such geometric structure underneath. Finally, various works have directly estimated the intrinsic dimensionality to be orders of magnitude smaller than their input dimension for diverse datasets, including images or physics data~\citep{levina_maximum_2004, bac_scikit-dimension_2021}.

\subsection*{How does GeoMatching behave when the dimensionality of the manifold rises?}

The difficulty of learning a manifold (i.e., number of samples required) scales exponentially with the intrinsic dimensionality of the manifold, polynomially on the curvature and linearly on the intrinsic volume of the manifold~\citep{narayanan_sample_2010}. Solving an ODE, needed to compute a single geodesic distance, scales linearly with the intrinsic dimensionality~\citep{kramer_probabilistic_2022}.

Thus, as the intrinsic dimensionality of the manifold rises, the most challenging step is manifold learning. While undesirable, we note that performance for other tasks such as classification also depend exponentially on the manifold dimensionality~\citep{narayanan_sample_2009}. The triumph of modern deep learning to solve these tasks in a wide range of datasets provides strong implicit justification for the manifold hypothesis~\citep{loaiza-ganem_deep_2024}.

\subsection*{How can model selection be used to choose from different manifold learning algorithms?}

Model selection can be done if we have access to treatment effects for a small subset of datapoints. This is the case for our real-world experiments, as described in Section~\ref{subsec:experimental-setup}. Most often, ground truth ATE/ITE are not available, so further assumptions are needed. Model selection can then be done according to a different criterion, e.g., based on the quality of the learned representation, reconstruction loss, etc. In general, model selection for causal inference methods remains challenging and an active research area in the community.

\subsection*{Is ATE computed on a test set?}

Yes, results are reported in a separate split. We separate the dataset randomly in two splits: the first one is to select hyperparameters, and the second split is to report final performance results.

\subsection*{How to avoid that the learned representation loses outcome information?}

This is a very interesting remark, in fact we are currently working on a future extension of GeoMatching that incorporates outcome information, lying in-between regression methods that predict model outcomes and methods that learn a subspace related to outcomes. This is an emerging research area, it is non-trivial and out-of-scope of the current submission.
GeoMatching falls in the bucket of methods that learn a latent representation solely based on pre-treatment
covariates (vast majority of matching methods out there). Matching without looking at the outcome is
desirable to avoid undesired biases and inadvertently induce selection bias, redirecting on the importance of
separating the ``design'' and ``analysis'' phases of a non-randomized study~\citep{rubin_for_2008}. This is a fundamental assumption within the potential outcome framework in causal inference~\citep{pearl_causal_2016}.

Not looking at the outcome incurs the risk of learning a latent space that is
uninformative/irrelevant w.r.t treatment response, which would not be very useful for TE estimation. Losing valuable outcome information is a potential problem for any matching method that only relies on pre-treatment covariates, including Euclidean, Mahalanobis, and Propensity Score Matching. 

Nothing in all these matching methods as well as GeoMatching explicitly prevents losing outcome information in the latent representation. Yet, all these methods are still useful because they rely on another fundamental, reasonable assumption: that patients similar in covariate space will tend to respond similarly, and that observed variability in covariate space is related/predictive of treatment response. Thus, by preserving information/structure present in the pre-treatment covariates, these methods indirectly preserve outcome information.

\subsection*{How to solve the ODE system for computation of geodesics?}

 We leverage the Stochman public library\footnote{Stochman Library: \url{https://github.com/MachineLearningLifeScience/stochman (https://
github.com/MachineLearningLifeScience/stochman}.} for computation of geodesics. For each geodesic, we solve an ODE numerically using the standard Runge-Kutta method~\citep{butcher_numerical_2016}.
We highlight that GeoMatching can be implemented with any choice of ODE solver underneath, benefiting from any advancement in the research front of ODE solvers.

\subsection*{How does the number of neighbors used for matching influence TE estimation?}

As the number of considered neighbors for matching increases, the TE estimation will incur in higher bias (towards average treatment response) but lower variance (less sensitivity to measurement noise in the outcome). These effects apply to any matching method, regardless of the considered distance~\citep{stuart_matching_2010}. In our work, we focus on the impact of improving the considered space (latent representation) and distance to perform matching.

\subsection*{If the data lies on a noisy manifold, are there methods to learn the denoised manifold? Can GeoMatching be applied in such situations?}

All experiments in the paper involve a noisy manifold to some extent. For example, in the synthetic 3D-swissroll from Section~\ref{subsec:swissroll}, we make the manifold noisy by adding $(D-3)$ dimensions of pure Gaussian noise. In the rest of scenarios, real covariates are unlikely to lie on a perfect manifold, but rather on a noisy one. The proposed strategy of PCA projection + fitting a parameterized Riemannian metric is empirically effective to recover the intrinsic denoised manifold. An alternative is to model the manifold noise explicitly using a probabilistic latent variable model like a GPLVM to learn a stochastic Riemannian manifold, but this approach is computationally more expensive and was harder to train in our experience. Of course, one can investigate methods to reconstruct the manifold that has different levels of tolerance and plug them into our pipeline, but this is beyond the scope of this work.

\subsection*{What happens when the covariates are not lying on a lower dimensional manifold? Are there ways to detect it?
}

If the covariates are not lying on a lower dimensional manifold, GeoMatching is equivalent to standard matching with Euclidean distance and should perform similarly. There exist several methods to estimate the intrinsic dimensionality of a manifold~\cite{levina_maximum_2004, bac_scikit-dimension_2021}. Interestingly, such works have estimated that the intrinsic dimensionality of high-dimensional data is generally orders of magnitude smaller than their input dimension.

\subsection*{If knowledge of the causal graph among the covariates is provided, can this be used to inform the construction of the latent space? If yes, can one combine GeoMatching with some causal discovery algorithm?}

Indeed, GeoMatching could be combined with causal discovery algorithms. In our introduction, we discuss how the causal mechanism across variables creates the manifold structure we use. Knowing this graph can help constrain manifold reconstruction. Instead of learning the entire manifold at once, one could learn all independent mechanisms separately and explicitly parameterize the manifold for computing geodesics, improving computational efficiency and accuracy. A causal discovery algorithm could integrate these distances with respect to the equivalence family of discovered graphs.
% We will mention this direction in our discussion. Also, we will elaborate more on the connection of this work with Dominguez-Olmedo et.al 2023 in the revised version of the manuscript.

\subsection*{Is it possible to have an isometry between the latent manifold and the outcome space if the latent manifold has a dimension larger than one?}

If the latent manifold has higher dimensionality than the output space, it is generally not possible to have an isometry. This is because reducing the dimensionality typically involves some loss of information, which means that the distances between points cannot be perfectly preserved. The latent manifold can nonetheless live in any space of higher dimensionality, which gives some degree of freedom to our proposed procedure.

In the 3D-swissroll example in Section~\ref{subsec:swissroll}, the 1D latent manifold exists in a 2D space. Because there exists an isometry between the latent manifold and the outcome space, GeoMatching is optimal. In this example, finding the latent manifold is easy, and GeoMatching provides a mechanism to predict the output function by computing Riemannian distances.

In general, the goal is to approximate the isometry by preserving as much of the relevant geometric structure as possible. Our proposed approach leverages the intuition that projecting the original datapoints into a distance-preserving manifold of lower dimensionality removes the extra covariates that would make predicting the output function directly more difficult. Given the projected data, we then compute Riemannian distances, which generalizes other distances such as Euclidean or Mahalanobis, resulting in similar performance if the learned projection is ``flat’’, and better performance if the learned projection is a curved manifold. With GeoMatching, a user does not need to worry about the shape of the manifold. We show this intuition holds in our other examples.

% \section{My Proof of Theorem 1}
% This is a boring technical proof.
% \section{My Proof of Theorem 2}
% This is a complete version of a proof sketched in the main text.

\end{document}